\documentclass[sigconf]{acmart}
\AtBeginDocument{%
  \providecommand\BibTeX{{%
    \normalfont B\kern-0.5em{\scshape i\kern-0.25em b}\kern-0.8em\TeX}}}
\usepackage{amsmath}
\usepackage{float} 
\usepackage{subfigure} 
\usepackage[ruled,linesnumbered]{algorithm2e}

\usepackage{amssymb}
\usepackage{amsbsy}
\usepackage{multirow}
 
\usepackage{wrapfig}
\usepackage{bbding}
\usepackage{color}
\usepackage{url}
\usepackage{fancyhdr}

\copyrightyear{2022}
\acmYear{2022}
\setcopyright{acmcopyright}\acmConference[WWW '22]{Proceedings of the ACM Web Conference 2022}{April 25--29, 2022}{Virtual Event, Lyon, France}
\acmBooktitle{Proceedings of the ACM Web Conference 2022 (WWW '22), April 25--29, 2022, Virtual Event, Lyon, France}
\acmPrice{15.00}
\acmDOI{10.1145/3485447.3512156}
\acmISBN{978-1-4503-9096-5/22/04}
\begin{document}
\title{SimGRACE: A Simple Framework for Graph Contrastive Learning without Data Augmentation}
\author{Jun Xia{$^{1,2\dagger}$}, Lirong Wu{$^{1,2\dagger}$}, Jintao Chen{$^{3}$}, Bozhen Hu{$^{1,2}$}, Stan Z.Li{$^{1,2\star}$}}
\affiliation{
  \institution{
   $^1$
  School of Engineering, Westlake University, Hangzhou 310030, China\\
   $^2$
   Institute of Advanced Technology, Westlake Institute for Advanced Study, Hangzhou 310030, China\\
  $^3$
  Zhejiang University, Hangzhou 310058, China\
  }
  \country{}
}

\email{{xiajun,wulirong, hubozhen,stan.zq.li}@westlake.edu.cn,chenjintao@zju.edu.cn}
\renewcommand{\authors}{Jun Xia, Lirong Wu, Jintao Chen, Bozhen Hu, Stan Z.Li}
\renewcommand{\shortauthors}{Jun Xia, et al.}
\begin{abstract}
Graph contrastive learning (GCL) has emerged as a dominant technique for graph representation learning which maximizes the mutual information between paired graph augmentations that share the same semantics. Unfortunately, it is difficult to preserve semantics well during augmentations in view of the diverse nature of graph data. Currently, data augmentations in GCL broadly fall into three unsatisfactory ways. First, the augmentations can be manually picked per dataset by trial-and-errors. Second, the augmentations can be selected via cumbersome search. Third, the augmentations can be obtained with expensive domain knowledge as guidance. All of these limit the efficiency and more general applicability of existing GCL methods. To circumvent these crucial issues, we propose a \underline{Sim}ple framework for \underline{GRA}ph \underline{C}ontrastive l\underline{E}arning, \textbf{SimGRACE} for brevity, which does not require data augmentations. Specifically, we take original graph as input and GNN model with its perturbed version as two encoders to obtain two correlated views for contrast. SimGRACE is inspired by the observation that graph data can preserve their semantics well during encoder perturbations while not requiring manual trial-and-errors, cumbersome search or expensive domain knowledge for augmentations selection. Also, we explain why SimGRACE can succeed. Furthermore, we devise adversarial training scheme, dubbed \textbf{AT-SimGRACE}, to enhance the robustness of graph contrastive learning and theoretically explain the reasons. Albeit simple, we show that SimGRACE can yield competitive or better performance compared with state-of-the-art methods in terms of generalizability, transferability and robustness, while enjoying unprecedented degree of flexibility and efficiency. The code is available at: \textcolor{magenta}{\url{https://github.com/junxia97/SimGRACE}}.
\end{abstract}
\thanks{$^{\dagger}$Equal Contribution, $^{\star}$Corresponding Author.}
\ccsdesc[500]{Computing methodologies~Neural networks}
\ccsdesc[500]{Computing methodologies~Learning latent representations}
\ccsdesc[500]{Mathematics of computing~Graph algorithms}
\keywords{Graph neural networks, graph self-supervised learning, contrastive learning, graph representation learning, robustness}
\maketitle
\section{Introduction}
\begin{figure}[t]
    \subfigure{
    \label{fig1-a}
   \includegraphics[width=0.15\textwidth]{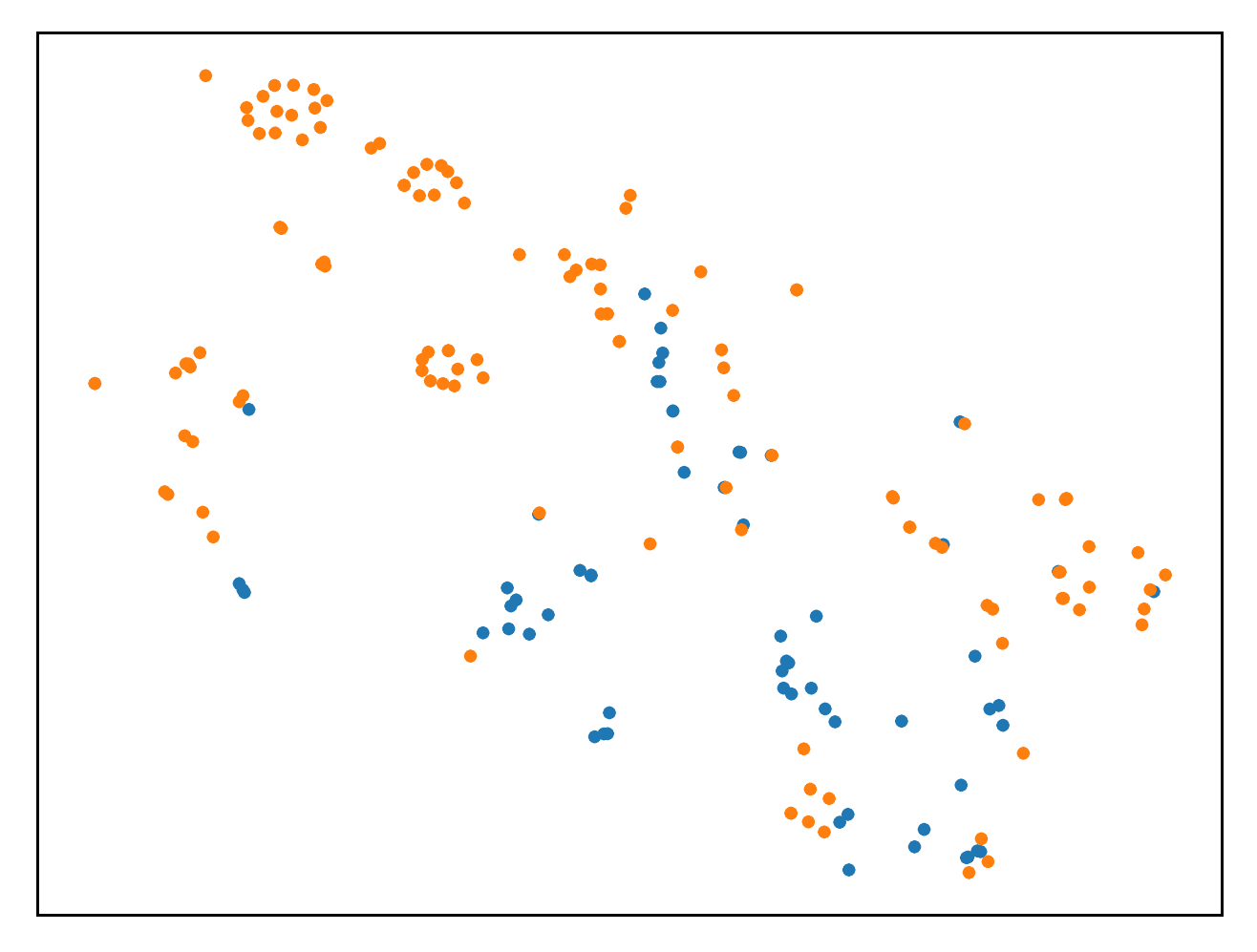}}
    \subfigure{
    \label{fig1-b}
    \includegraphics[width=0.15\textwidth]{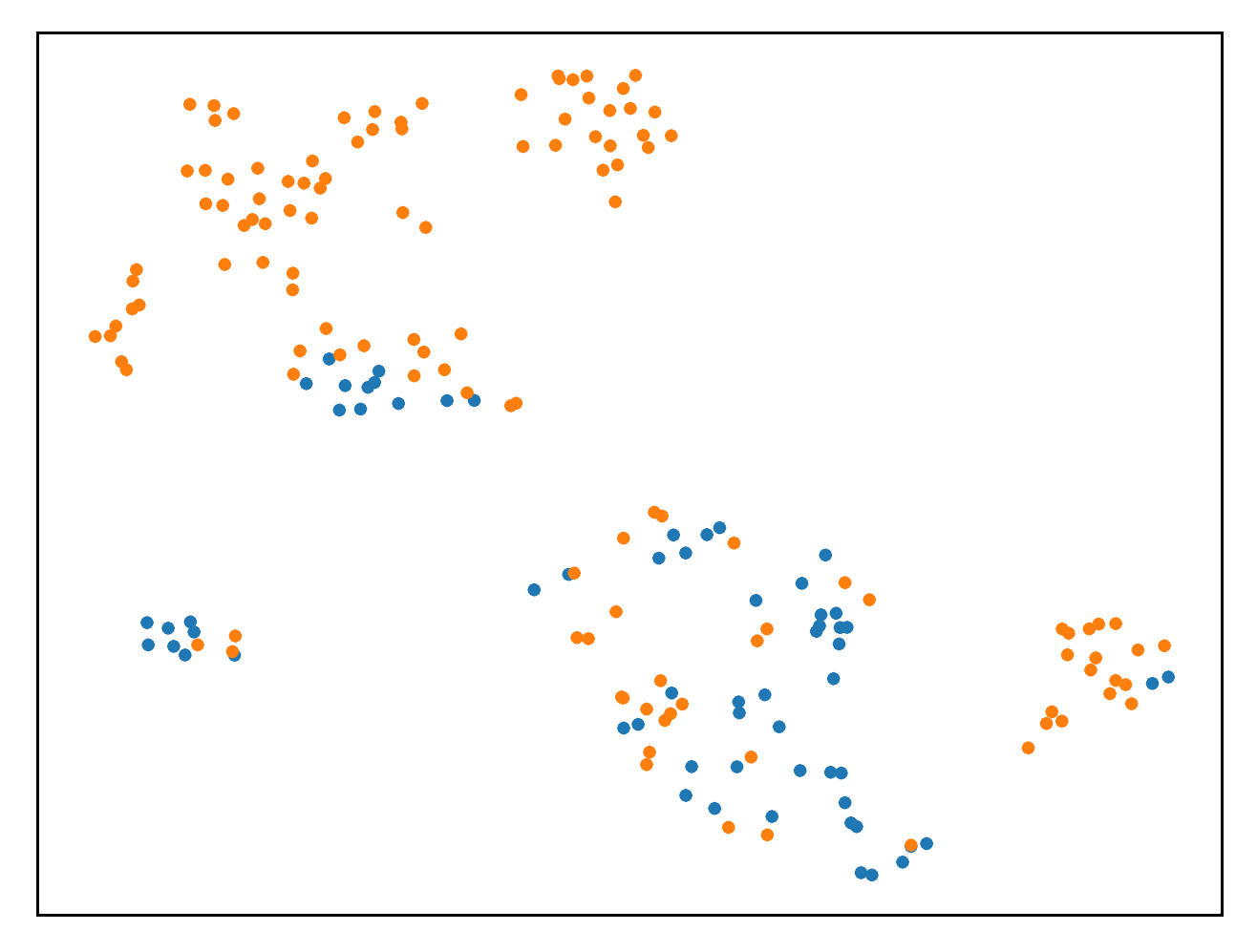}}
    \subfigure{
    \label{fig1-c}
    \includegraphics[width=0.15\textwidth]{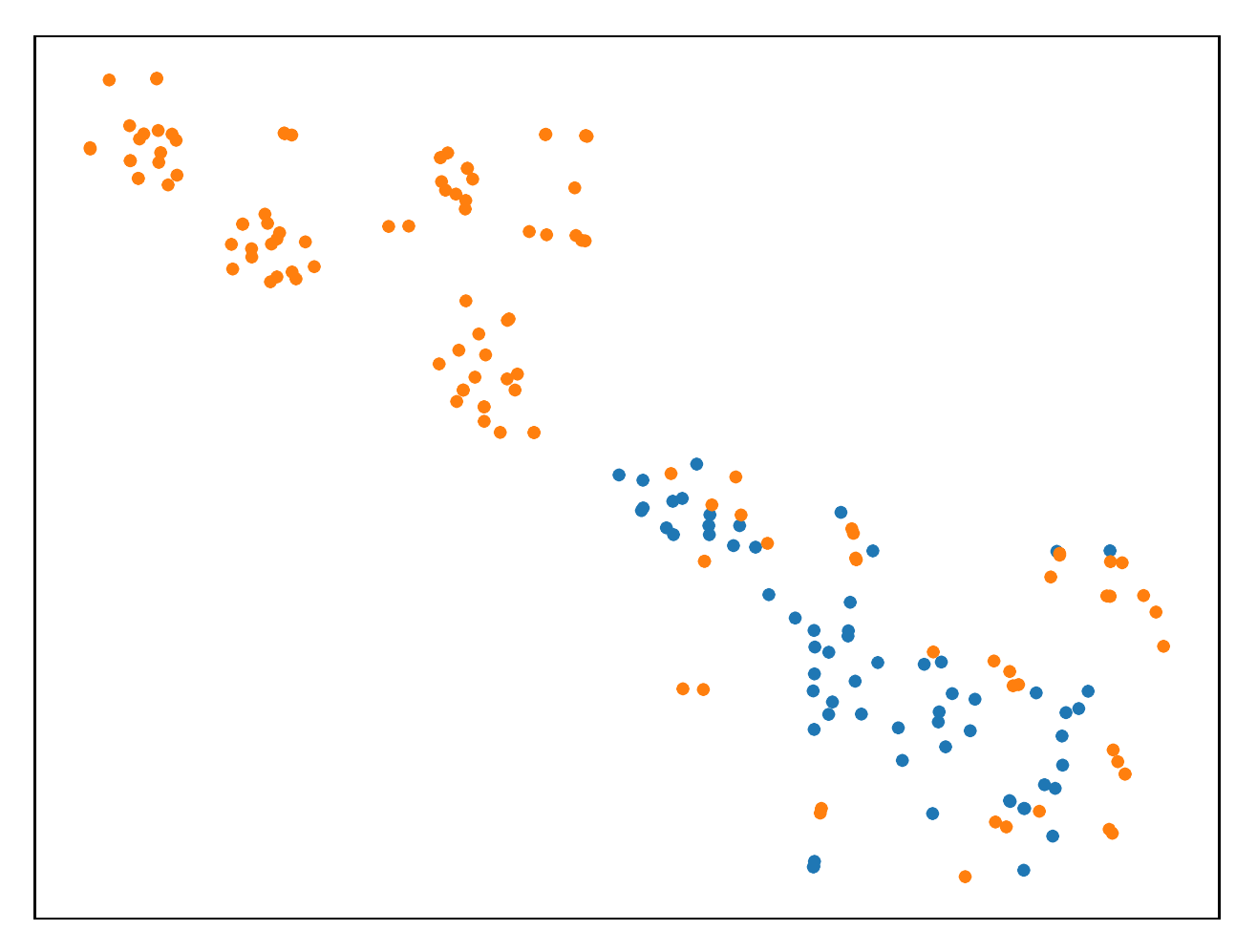}}
    \subfigure{
    \label{fig1-d}
    \includegraphics[width=0.15\textwidth]{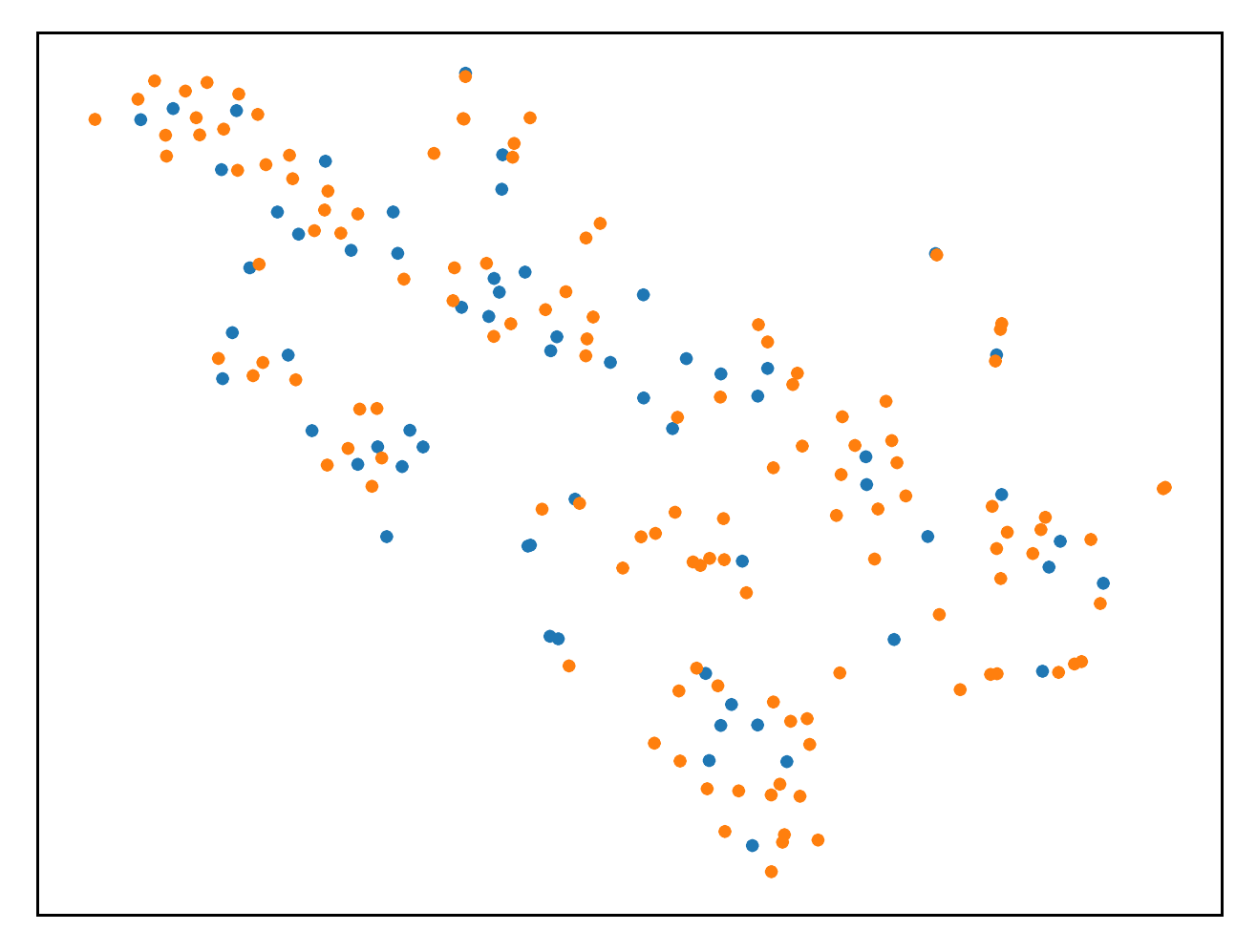}}
    \subfigure{
    \label{fig1-e}
    \includegraphics[width=0.15\textwidth]{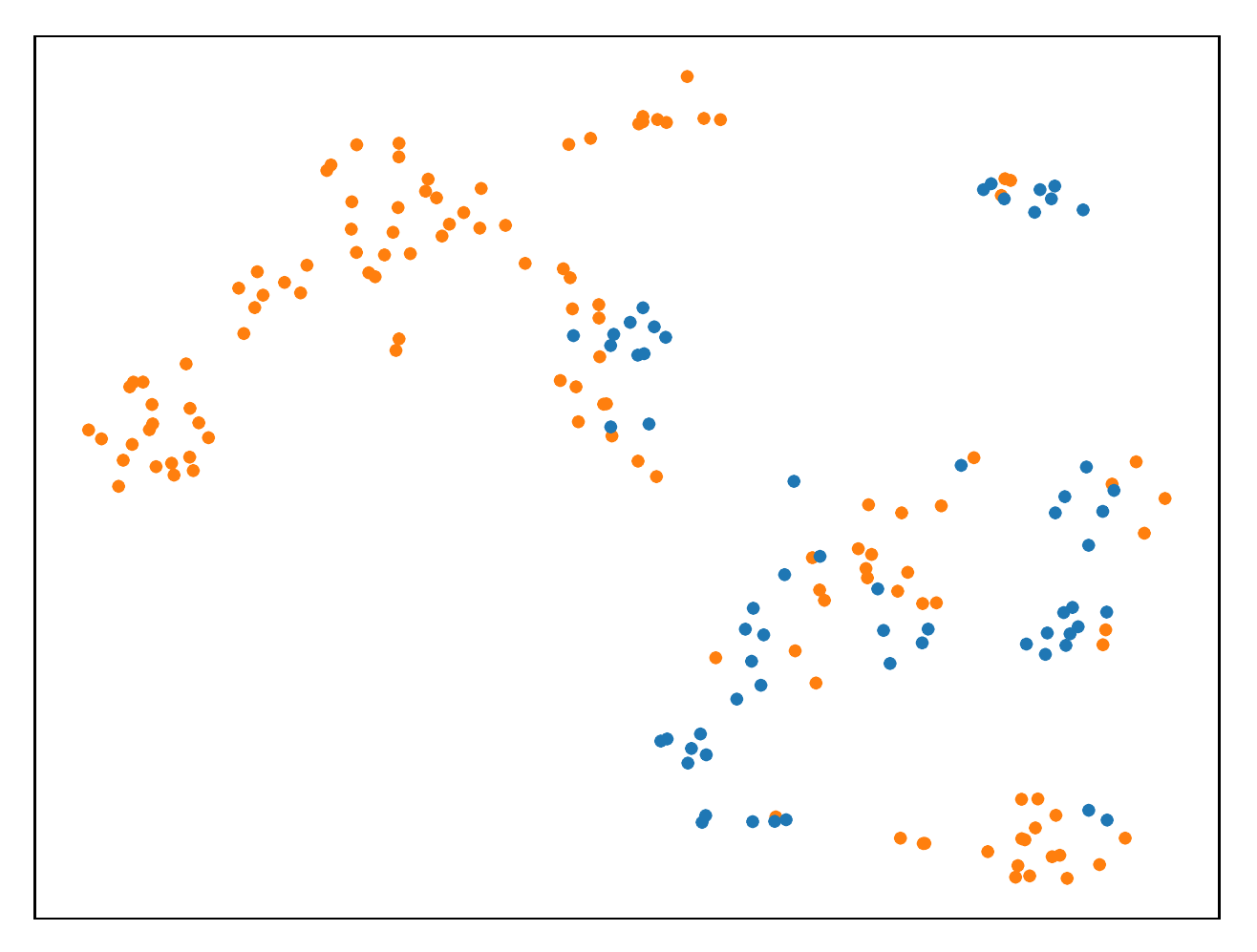}}
    \subfigure{
    \label{fig1-f}
    \includegraphics[width=0.15\textwidth]{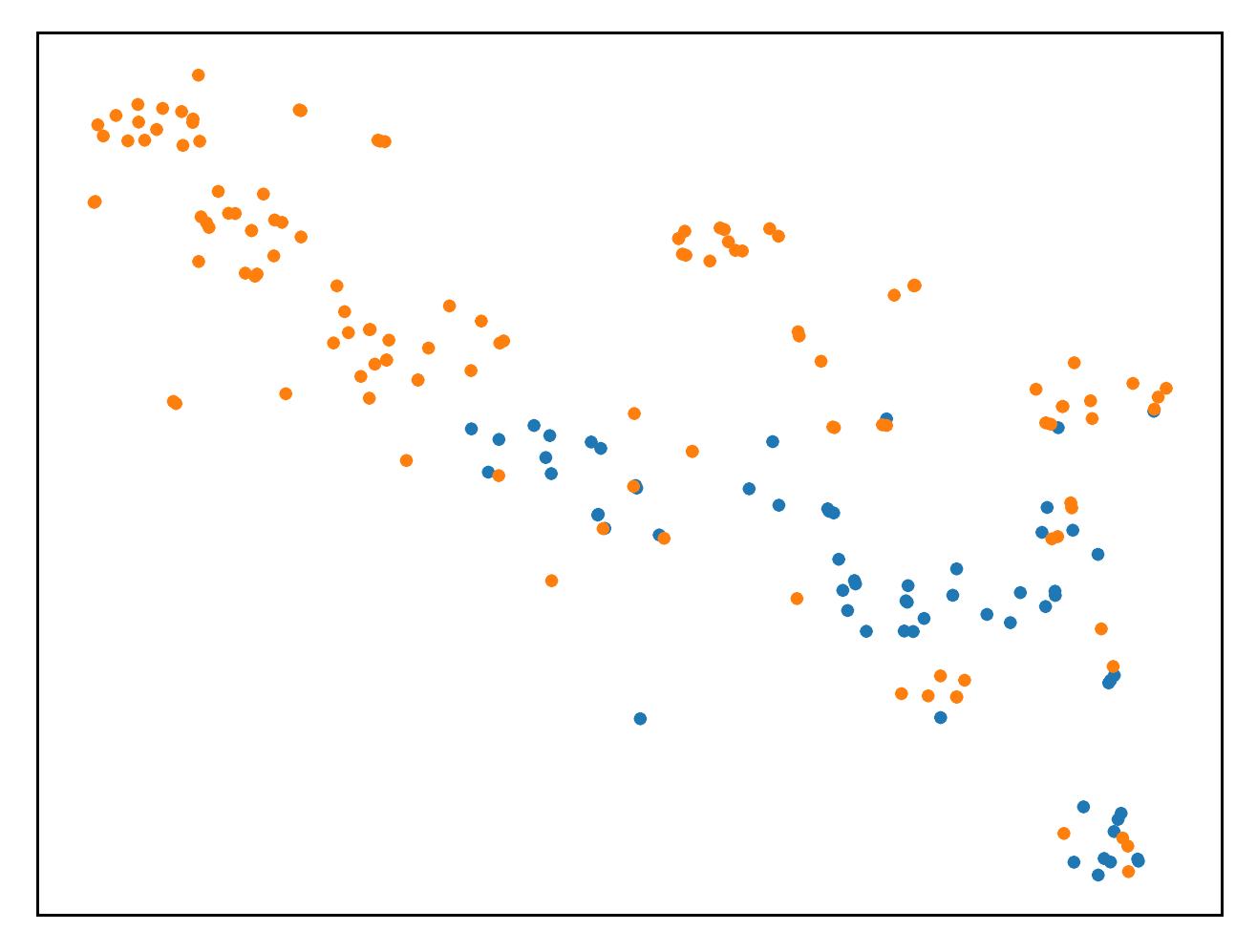}}
   \leftline{\qquad \quad GraphCL \qquad \qquad \quad MoCL \qquad \qquad \qquad SimGRACE} 

    \caption{Comparison of GraphCL~\cite{You2020GraphCL}, MoCL~\cite{sun2021mocl} and SimGRACE on MUTAG dataset. The samples of two classes are distinguished by colors (blue \& orange). We first train three GNN encoders with these methods respectively and visualise the representations of original graphs with t-SNE in the upper row. Then, we perturb graphs or encoders in their respective ways (edge perturbation for GraphCL, replacing functional group with bioisosteres of similar properties for MoCL, encoder perturbation for SimGRACE) and visualise the representations of perturbed (GraphCL, MoCL) or original (SimGRACE) graphs in the below row. Unlike GraphCL, SimGRACE and MoCL can preserve the class identity semantics well after perturbations. However, MoCL requires expensive domain knowledge as guidance.}
    \label{fig1}
\end{figure}
Graph Neural Networks (GNNs), inheriting the power of neural networks and utilizing the structural information of graph data simultaneously, have achieved overwhelming accomplishments in various graph-based tasks, such as node, graph classification or graph generation~\cite{kipf2016semi,xu2018how,du2021graphgt}. However, most existing GNNs are trained in a supervised manner and it is often resource- and time-intensive to collect abundant true-labeled data~\cite{xia2021towards,tan2021co,xia2022otcleaner}. To remedy this issue, tremendous endeavors have been devoted to graph self-supervised learning that learns representations from unlabeled  graphs. Among many, graph contrastive learning (GCL)~\cite{xia2022survey,You2020GraphCL,you2021graph} follows the general framework of contrastive learning in computer vision domain~\cite{ting2020a,wu2018unsupervised}, in which two augmentations are generated for each graph and then maximizes the mutual information between these two augmented views. In this way, the model can learn representations that are invariant to perturbations. For example, GraphCL~\cite{You2020GraphCL} first designs four types of general augmentations (node dropping, edge perturbation, attribute masking and subgraph) for GCL. However, these augmentations are not suitable for all scenarios because the structural information and semantics of the graphs varies significantly across domains. For example, GraphCL~\cite{You2020GraphCL} finds that edge perturbation benefits social networks but hurt some biochemical molecules in GCL. Worse still, these augmentations may alter the graph semantics completely even if the perturbation is weak. For example, dropping a carbon atom in the phenyl ring will alter the aromatic system and result in an alkene chain, which will drastically change the molecular properties~\cite{sun2021mocl}. 

\begin{table*}[t]
\caption{Comparison between state-of-the-art GCL methods (graph-level representation learning) and SimGRACE.}
\label{Table intro1}
\centering
\small
\begin{tabular}{@{}cccccc@{}}
\toprule
        & No manual trial-and-errors & No domain knowledge & Preserving semantics & No cumbersome search & Generality \\ \midrule
GraphCL~\cite{You2020GraphCL} & \XSolidBrush           & \Checkmark                    & \XSolidBrush                     & \Checkmark          & \XSolidBrush           \\
MoCL~\cite{sun2021mocl}    & \Checkmark           & \XSolidBrush        &  \Checkmark            &  \Checkmark  & \XSolidBrush                  \\
JOAO(v2)~\cite{you2021graph}    & \Checkmark           & \Checkmark                    & \XSolidBrush                     & \XSolidBrush    & \Checkmark                  \\
SimGRACE  & \Checkmark &\Checkmark                     & \Checkmark                     & \Checkmark      & \Checkmark                \\ \bottomrule
\end{tabular}
\end{table*}
To remedy these issues, several strategies have been proposed recently. Typically, GraphCL~\cite{You2020GraphCL} manually picks data augmentations per dataset by tedious trial-and-errors, which significantly limits the generality and practicality of their proposed framework. To get rid of the tedious dataset-specific manual tuning of GraphCL, JOAO~\cite{you2021graph} proposes to automate GraphCL in selecting augmentation pairs. However, it suffers more computational overhead to search suitable augmentations and still relies on human prior knowledge in constructing and configuring the augmentation pool to select from. To avoid altering the semantics in the general augmentations adopted in GraphCL and JOAO(v2), MoCL~\cite{sun2021mocl} proposes to replace valid substructures in molecular graph with bioisosteres that share similar properties. However, it requires expensive domain knowledge as guidance and can not be applied in other domains like social graphs. Hence, a natural question emerges: \emph{Can we emancipate graph contrastive learning from tedious manual trial-and-errors, cumbersome search or expensive domain knowledge ?} \\
To answer this question, instead of devising more advanced data augmentations strategies for GCL, we attempt to break through state-of-the-arts GCL framework which takes semantic-preserved data augmentations as prerequisite. More specifically, we take original graph data as input and GNN model with its perturbed version as two encoders to obtain two correlated views. And then, we maximize the agreement of these two views. With the encoder perturbation as noise, we can obtain two different embeddings for same input as “positive pairs”. Similar to previous works~\cite{ting2020a,You2020GraphCL}, we take other graph data in the same mini-batch as “negative pairs”. The idea of encoder perturbation is inspired by the observations in Figure~\ref{fig1}. The augmentation or perturbation of MoCL and our SimGRACE can preserve the class identity semantics well while GraphCL can not. Also, we explain why SimGRACE can succeed. Besides, GraphCL~\cite{You2020GraphCL} shows that GNNs can gain robustness using their proposed framework. However, (1) they do not explain why GraphCL can enhance the robustness; (2) GraphCL seems to be immunized to random attacks well while performing unsatisfactory against adversarial attacks. GROC~\cite{jovanovic2021towards} first integrates adversarial transformations into the graph contrastive learning framework and improves the robustness against adversarial attacks. Unfortunately, as the authors pointed out, the robustness of GROC comes at a price of much longer training time because conducting adversarial transformations for each graph is time-consuming. 
To remedy these deficiencies, we propose a novel algorithm AT-SimGRACE to perturb the encoder in an adversarial way, which introduces less computational overhead while showing better robustness. Theoretically, we explain why AT-SimGRACE can enhance the robustness. We highlight our contributions as follows:
\begin{itemize}
    \item[$\bullet$] \emph{Significance:} We emancipate graph contrastive learning from tedious manual trial-and-errors, cumbersome search or expensive domain knowledge which limit the efficiency and more general applicability of existing GCL methods. The comparison between SimGRACE and state-of-the-art GCL methods can be seen in Table~\ref{Table intro1}.
    \item[$\bullet$] \emph{Framework:} We develop a novel and effective framework, SimGRACE, for graph contrastive learning which enjoys unprecedented degree of flexibility, high efficiency and ease of use. Moreover, we explain why SimGRACE can succeed.
    \item[$\bullet$] \emph{Algorithm:} We propose a novel algorithm AT-SimGRACE to enhance the robustness of graph contrastive learning. AT-SimGRACE can achieve better robustness while introducing minor computational overhead. 
    \item[$\bullet$] \emph{Experiments:} We experimentally show that the proposed methods can yield competitive or better performance compared with state-of-the-art methods in terms of generalizability, transferability, robustness and efficiency on multiple social and biochemical graph datasets.
\end{itemize}
\begin{figure*}[t]
    \begin{center}
    \includegraphics[width=0.65\textwidth]{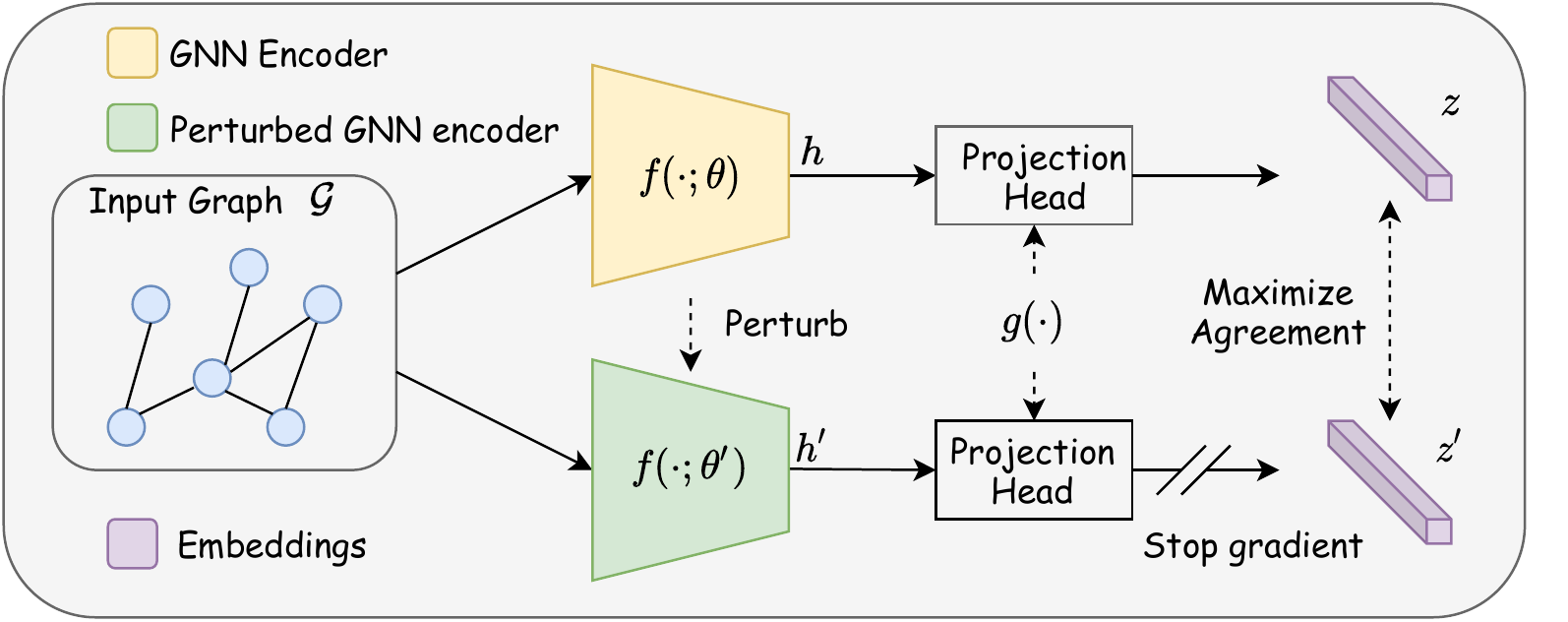}
    \end{center}
    \caption{Illustration of SimGRACE, a simple framework of graph contrastive learning. Instead of augmenting the graph data, we feed the original graph $\mathcal{G}$ into a GNN encoder $f(\cdot;\boldsymbol{\theta})$ and its perturbed version $f(\cdot;\boldsymbol{\theta}^{\prime})$. After passing a shared projection
head $g(\cdot)$, we maximize the agreement between representations $\boldsymbol{z}_i$ and $\boldsymbol{z}_j$ via a contrastive loss.}
    \label{fig_2}
\end{figure*}
\section{Related work}
\label{gen_inst}
\subsection{Generative / Predictive self-supervised learning on graphs}
Inspired by the success of self-supervised learning in computer vision~\cite{kaiming2020momentum,ting2020a} and natural language processing~\cite{devlin2019bert,lan2020albert,zheng2022using}, tremendous endeavors have been devoted to graph self-supervised learning that learns representations in an unsupervised manner with designed pretext tasks. Initially, Hu et al.~\cite{hu2020strategies} propose two pretext tasks, i.e, predicting neighborhood context and node attributes to conduct node-level pre-training. Besides, they utilize supervised graph-level property prediction and structure similarity prediction as pretext tasks to perform graph-level pre-training. GPT-GNN~\cite{hu2020gpt-gnn} designs generative task in which node attributes and edges are alternatively generated such that the likelihood of a graph is maximized. Recently, GROVER~\cite{rong2020self-supervised} incorporates GNN into a transformer-style architecture and learns node embedding by predicting contextual property and graph-level motifs. We recommend the readers to refer to a recent survey~\cite{xia2022survey} for more information. Different from above methods, our SimGRACE follows a contrastive framework that will be introduced below.

\subsection{Graph Contrastive Learning}
Graph contrastive learning can be categorized into two groups. One group can encode useful information by contrasting local and global representations. Initially, DGI~\cite{velickovic2019deep} and InfoGraph~\cite{sun2020infograph} are proposed to obtain expressive representations for graphs or nodes via maximizing the mutual information between graph-level representations and substructure-level representations of different granularity. More recently, MVGRL~\cite{hassani2020contrastive} proposes to learn both node-level and graph-level representation by performing node diffusion and contrasting node representation to augmented graph representations. Another group is designed to learn representations that are tolerant to data transformation. Specifically, they first augment graph data and feed the augmented graphs into a shared encoder and projection head, after which their mutual information is maximized. Typically, for node-level tasks~\cite{Zhu:2021tu, Zhu:2020vf}, GCA~\cite{zhu2021graph} argues that data augmentation schemes should preserve intrinsic structures and attributes of graphs and thus proposes to adopt adaptive augmentations that only perturb unimportant components. DGCL~\cite{DBLP:journals/corr/abs-2110-02027} introduces a novel probabilistic method to alleviate the issue of false negatives in GCL. For graph-level tasks, GraphCL~\cite{You2020GraphCL} proposes four types of augmentations for general graphs and demonstrated that the learned representations can help downstream tasks. However, the success of GraphCL comes at the price of tedious manual trial-and-errors. To tackle this issue, JOAO~\cite{you2021graph} proposes a unified bi-level optimization framework to automatically select data augmentations for GraphCL, which is time-consuming and inconvenient. More recently, MoCL~\cite{sun2021mocl} proposes to incorporate domain knowledge into molecular graph augmentations in order to preserve the semantics. However, the domain knowledge is extremely expensive. Worse still, MoCL can only work on molecular graph data, which significantly limits their generality. Despite the fruitful progress, they still require tedious manual trial-and-errors, cumbersome search or expensive domain knowledge for augmentation selection. Instead, our SimGRACE breaks through state-of-the-arts GCL framework that takes semantic-preserved data augmentations as prerequisite.
\section{Method}
\subsection{SimGRACE}
In this section, we will introduce SimGRACE framework in details. As sketched in Figure~\ref{fig_2}, the framework consists of the following three major components:

\textbf{(1) Encoder perturbation.} A GNN encoder $f(\cdot;\boldsymbol{\theta})$ and its its perturbed version $f(\cdot;\boldsymbol{\theta}^{\prime})$ first extract two graph-level representations $\mathbf{h}$ and $\mathbf{h}^{\prime}$ for the same graph $\mathcal{G}$, which can be formulated as, 
\begin{gather}
\mathbf{h}=f(\mathcal{G};\boldsymbol{\theta}), \mathbf{h}^{\prime}=f(\mathcal{G};\boldsymbol{\theta}^{\prime}).
\end{gather}
The method we proposed to perturb the encoder $f(\cdot;\boldsymbol{\theta})$ can be mathematically described as,
\begin{equation}
\boldsymbol{\theta}^{\prime}_{l}=\boldsymbol{\theta}_{l}+\eta \cdot \boldsymbol{\Delta}\boldsymbol{\theta}_{l} ; \quad \boldsymbol{\Delta}\boldsymbol{\theta}_{l} \sim \mathcal{N}\left(0, \sigma_{l}^{2}\right),
\end{equation}
where $\boldsymbol{\theta}_{l}$ and $\boldsymbol{\theta}^{\prime}_{l}$ are the weight tensors of the $l$-th layer of the GNN encoder and its perturbed version respectively. $\eta$ is the coefficient that scales the magnitude of the perturbation. $\boldsymbol{\Delta}\boldsymbol{\theta}_{l}$ is the perturbation term which samples from Gaussian distribution with zero mean and variance $\sigma_{l}^{2}$. Also, we show that the performance will deteriorate when we set $\eta=0$ in section~\ref{HP}. Note that BGRL~\cite{thakoor2021bootstrapped} and MERIT~\cite{jin2021multi-scale} also update a target network with an online encoder during training. However, SimGRACE differs from them in three aspects: (1) SimGRACE perturbs the encoder with a random Guassian noise instead of momentum updating; (2) SimGRACE does not require data augmentation while BGRL and MERIT take it as prerequisite. (3) SimGRACE focuses on graph-level representation learning while BGRL and MERIT only work in node-level tasks.

\textbf{(2) Projection head.} As advocated in~\cite{ting2020a}, a non-linear transformation $g(\cdot)$ named projection head maps the representations to another latent space can enhance the performance. In our SimGRACE framework, we also adopt a two-layer perceptron (MLP) to obtain $z$ and $z^{\prime}$,
\begin{gather}
z=g(\mathbf{h}), z^{\prime}=g(\mathbf{h}^{\prime}).
\end{gather}

\textbf{(3) Contrastive loss.} In SimGRACE framework, we utilize the normalized temperature-scaled cross entropy loss (NT-Xent) as previous works~\cite{sohn2016improved,oord2019representation,wu2018unsupervised,You2020GraphCL} to enforce the agreement between positive pairs $z$ and $z^{\prime}$ compared with negative pairs.

During SimGRACE training, a minibatch of $N$ graphs are randomly sampled and then they are fed into a GNN encoder $f(\cdot;\boldsymbol{\theta})$ and its perturbed version $f(\cdot;\boldsymbol{\theta}^{\prime})$, resulting in two presentations for each graph and thus $2N$ representations in total. We re-denote $z, z^{\prime}$ as $\boldsymbol{z}_n, \boldsymbol{z}_{n}^{\prime}$ for $n$-th graph in the minibatch. Negative pairs are generated from the other $N-1$ perturbed representations within the same mini-batch as in~\cite{chen2017on,ting2020a,You2020GraphCL}. Denoting the cosine similarity function as $\operatorname{sim}\left(\boldsymbol{z}, \boldsymbol{z}^{\prime}\right)=\boldsymbol{z}^{\top} \boldsymbol{z}^{\prime} /\left\|\boldsymbol{z}\right\|\left\|\boldsymbol{z}^{\prime}\right\|$, the contrastive loss for the $n$-th graph is defined as,
\begin{gather}
\label{con}
\ell_{n}=-\log \frac{\exp \left(\operatorname{sim}\left(\boldsymbol{z}_{n}, \boldsymbol{z}^{\prime}_{n}\right)) / \tau\right)}{\sum_{n^{\prime}=1, n^{\prime} \neq n}^{N} \exp \left(\operatorname{sim}\left(\boldsymbol{z}_{n}, \boldsymbol{z}_{n^{\prime}}\right) / \tau\right)},
\end{gather}
where $\tau$ is the temperature parameter. The final loss is computed across all positive pairs in the minibatch. 

\subsection{Why can SimGRACE work well?}
In order to understand why SimGRACE can work well, we first introduce the analysis tools from~\cite{tongzhou2020understanding}. Specifically, they identify two key properties related to contrastive learning: \emph{alignment} and \emph{uniformity} and then
propose two metrics to measure the quality of representations obtained via contrastive learning. One is the alignment metric which is straightforwardly defined with the expected distance between positive pairs:
\begin{equation}
\ell_{\text{align}}(f ; \alpha) \triangleq \underset{(x, y) \sim p_{\text {pos }}}{\mathbb{E}}\left[\|f(x)-f(y)\|_{2}^{\alpha}\right], \quad \alpha>0
\end{equation}
where $p_{\text{pos}}$ is the distribution of positive pairs (augmentations of the same sample). This metric is well aligned with the objective of contrastive learning: positive samples should
stay close in the embedding space. Analogously, for our SimGRACE framework, we provide a modified metric for \emph{alignment},
\begin{equation}
\ell_{\text{align}}(f ; \alpha) \triangleq \underset{x \sim p_{\text {data}}}{\mathbb{E}}\left[\|f(x;\boldsymbol{\theta})-f(x;\boldsymbol{\theta}^{\prime})\|_{2}^{\alpha}\right], \quad \alpha>0
\end{equation}
where $p_{\text{data}}$ is the data distribution. We set $\alpha=2$ in our experiments. The other is the uniformity metric which is defined as the logarithm of the average pairwise Gaussian potential:
\begin{equation}
\ell_{\text{uniform}}(f ; \alpha) \triangleq \log \underset{x, y \overset{i.i.d.}{\sim} p_{\text {data }}}{\mathbb{E}}\left[e^{-t\|f(x; \boldsymbol{\theta})-f(y; \boldsymbol{\theta})\|_{2}^{2}}\right]. \quad t>0
\end{equation}
In our experiments, we set $t=2$. The uniformity metric is also aligned with the objective of contrastive learning that the embeddings of random samples should scatter on the hypersphere. 
\begin{figure}[t]
    \centering
    \includegraphics[width=0.48\textwidth]{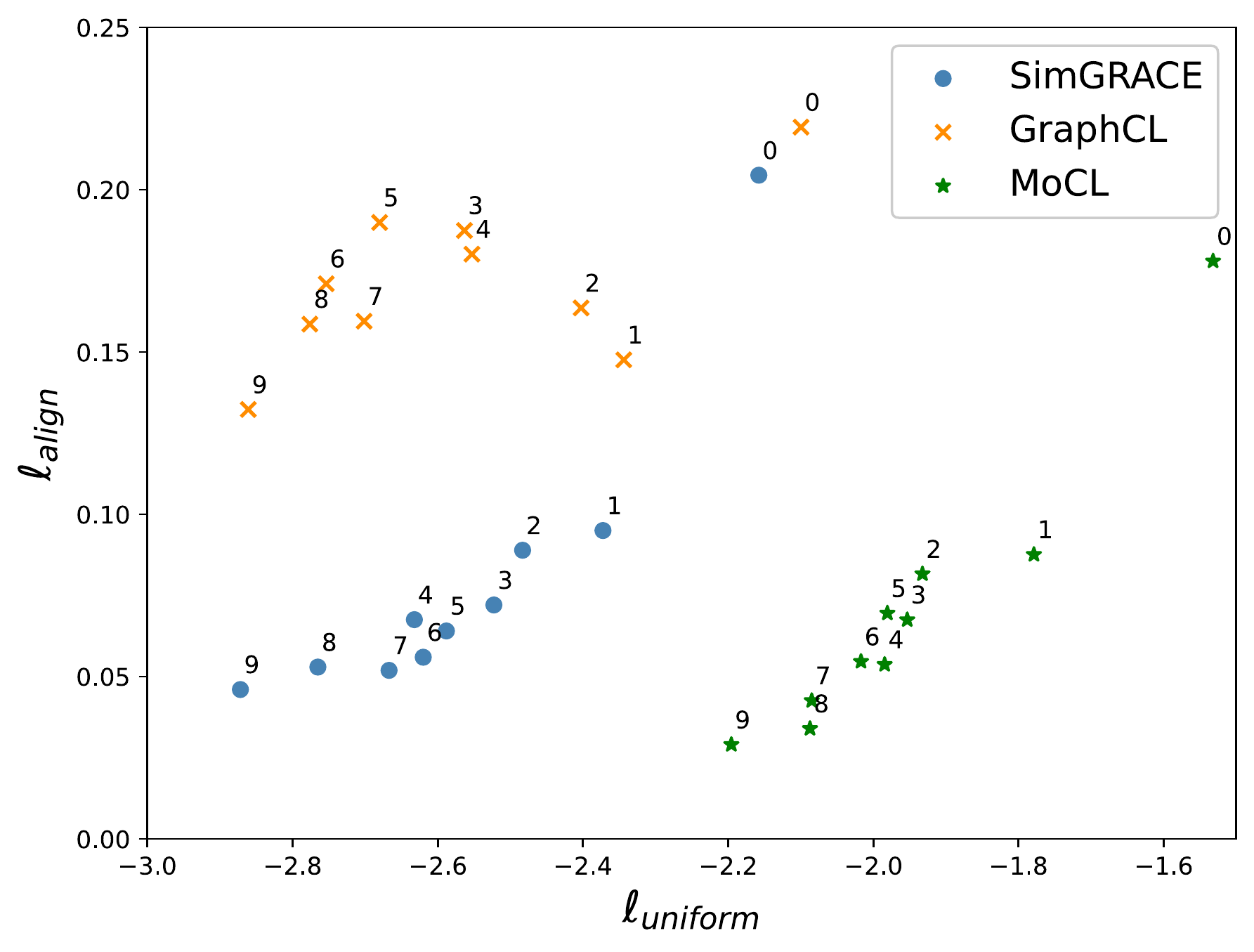}
   \Description[A bell-like histogram]{$\ell_{align}$-$\ell_{uniform}$ plot for SimGRACE, GraphCL and MoCL on MUTAG dataset. The numbers around the points are the indexes of epochs. For both $\ell_{align}$ and $\ell_{uniform}$, lower is better.}
   \caption{$\ell_{align}$-$\ell_{uniform}$ plot for SimGRACE, GraphCL and MoCL on MUTAG dataset. The numbers around the points are the indexes of epochs. For both $\ell_{align}$ and $\ell_{uniform}$, lower is better.}
   \label{fig3}
\end{figure}
We take the checkpoints of SimGRACE, GraphCL and MoCL every 2 epochs during training and visualize the alignment $\ell_{align}$ and uniformity $\ell_{uniform}$ metrics in Figure~\ref{fig3}. As can be observed, all the three methods can improve the alignment and uniformity. However, GraphCL achieves a smaller gain on the alignment than SimGRACE and MoCL. In other words, the positive pairs can not stay close in GraphCL because general graph data augmentations (drop edges, drop nodes and etc.) destroy the semantics of original graph data, which degrades the quality of the representations learned by GraphCL. Instead, MoCL augments graph data with domain knowledge as guidance and thus can preserve semantics during augmentation. Eventually, MoCL dramatically improves the alignment. Compared with GraphCL, SimGRACE can achieve better alignment while improving uniformity because encoder perturbation can preserve data semantics well. On the other hand, although MoCL achieves better alignment than SimGRACE via introducing domain knowledge as guidance, it only achieves a small gain on the uniformity, and eventually underperforms SimGRACE.  
\subsection{AT-SimGRACE}
Recently, GraphCL~\cite{You2020GraphCL} shows that GNNs can gain robustness using their proposed framework. However, they did not explain why GraphCL can enhance the robustness. Additionally, GraphCL seems to be immunized to random attacks well while being unsatisfactory against adversarial attacks. In this section, we aim to utilize Adversarial Training (AT)~\cite{goodfellow2015explaining,madry2019towards} to improve the adversarial robustness of SimGRACE in a principled way. Generally, AT directly incorporates adversarial examples into the training process to solve the following optimization problem:
\begin{equation}
\label{AT}
\min_{\boldsymbol{\theta}} \mathcal{L}^{\prime}(\boldsymbol{\theta}), \quad \text{where} \quad \mathcal{L}^{\prime}(\boldsymbol{\theta})=\frac{1}{n} \sum_{i=1}^{n} \max _{\left\|\mathbf{x}_{i}^{\prime}-\mathbf{x}_{i}\right\|_{p} \leq \epsilon} \ell_i^{\prime}\left(f\left(\mathbf{x}_{i}^{\prime};\boldsymbol{\theta}\right), y_{i}\right),
\end{equation}
where $n$ is the number of training examples, $\mathrm{x}_{i}^{\prime}$ is the adversarial example within the $\epsilon$-ball (bounded by an $L_{p}$-norm) centered at natural example $\mathbf{x}_{i}, f$ is the DNN with weight $\boldsymbol{\theta}, \ell^{\prime}(\cdot)$ is the standard supervised classification loss (e.g., the cross-entropy loss), and $\mathcal{L}^{\prime}(\boldsymbol{\theta})$ is called the "adversarial loss". However, above general framework of AT can not directly be applied in graph contrastive learning because (1) AT requires labels as supervision while labels are not available in graph contrastive learning; (2) Perturbing each graph for the dataset in an adversarial way will introduce heavy computational overhead, which has been pointed out in GROC~\cite{jovanovic2021towards}. To remedy the first issue, we substitute supervised classification loss in Eq. (\ref{AT}) with contrastive loss in Eq. (\ref{con}). To tackle the second issue, instead of conducting adversarial transformation of graph data, we perturb the encoder in an adversarial way, which is more computationally efficient.\\
Assuming that $\boldsymbol{\Theta}$ is the weight space of GNNs, for any $\mathbf{w}$ and any positive $\epsilon$, we can define the norm ball in $\boldsymbol{\theta}$ with radius $\epsilon$ centered at $\mathbf{w}$ as,
\begin{equation}
\mathbf{R}(\mathbf{w} ; \epsilon):=\{\boldsymbol{\boldsymbol{\theta}} \in \boldsymbol{\Theta}:\|\boldsymbol{\boldsymbol{\theta}}-\mathbf{w}\| \leq \epsilon\},
\end{equation}
we choose $L_{2}$ norm to define the norm ball in our experiments. With this definition, we can now formulate our AT-SimGRACE as an optimization problem, 
\begin{equation}
\begin{aligned}
\min_{\boldsymbol{\theta}} & \quad\mathcal{L}(\boldsymbol{\theta} + \boldsymbol{\Delta}), \\ \text{where} \quad \mathcal{L}(\boldsymbol{\theta} + \boldsymbol{\Delta})=\frac{1}{M} \sum_{i=1}^{M} &  \max_{\boldsymbol{\Delta}\in\mathbf{R}(\mathbf{0};\epsilon)} \ell_i\left(f\left(\mathcal{G}_{i};\boldsymbol{\theta}+\boldsymbol{\Delta}\right), f\left(\mathcal{G}_{i};\boldsymbol{\theta}\right)\right),
\end{aligned}
\end{equation}
where $M$ is the number of graphs in the dataset. We propose Algorithm~\ref{algorithm} to solve this optimization problem. Specifically, for inner maximization, we forward $I$ steps to update $\boldsymbol{\Delta}$ in the direction of increasing the contrastive loss using gradient ascent algorithm. With the output perturbation $\boldsymbol{\Delta}$ of inner maximization, the outer loops update the weights $\boldsymbol{\theta}$ of GNNs with mini-batched SGD.
\begin{algorithm}[ht]
\caption{Encoder perturbation of AT-SimGRACE}\label{algorithm}
\KwData{Graph dataset $\mathcal{D}=\left\{\mathcal{G}_1, \mathcal{G}_2,..., \mathcal{G}_M\right\}$, contrastive loss $\ell$, batch size $N$, initial encoder weights $\boldsymbol{\theta}$, inner iterations $I$, inner learning rate $\zeta$, outer learning rate $\gamma$, norm ball radius $\epsilon$.}
\For{each mini-batch}{
Sample $\mathcal{D}_B= \left\{\mathcal{G}_{i}\right\}_{i=1}^{N}$ from $\mathcal{D}$;\\
Initialize perturbation: $\boldsymbol{\Delta} \leftarrow 0$;\\
\For {$t = 0,1,2,...,I-1$}{
Update perturbation: $\boldsymbol{\Delta} \leftarrow \boldsymbol{\Delta} + \zeta\sum_{i=1}^{N} \nabla_{\boldsymbol{\boldsymbol{\theta}}} \ell_i\left(f\left(\mathcal{G}_{i};\boldsymbol{\theta}+\boldsymbol{\Delta}\right), f\left(\mathcal{G}_{i};\boldsymbol{\theta}\right)\right) / N$;\\
\If{$\left\|\boldsymbol{\Delta}\right\|_{2}>\epsilon$}{Normalize perturbation: $\boldsymbol{\Delta} \leftarrow \epsilon \boldsymbol{\Delta} /\left\|\boldsymbol{\Delta}\right\|_{2}$;\\}}
Update weights: $\boldsymbol{\boldsymbol{\theta}}^{\prime} \leftarrow \boldsymbol{\boldsymbol{\theta}}-\gamma \sum_{i=1}^{N} \nabla_{\boldsymbol{\boldsymbol{\theta}}} \ell_i\left(f\left(\mathcal{G}_{i};\boldsymbol{\theta}+\boldsymbol{\Delta}\right), f\left(\mathcal{G}_{i};\boldsymbol{\theta}\right)\right) / N.$
}
\end{algorithm}
\subsection{Theoretical Justification}
In this section, we aim to explain the reasons why AT-SimGRACE can enhance the robustness of graph contrastive learning. To start, it is widely accepted that flatter loss landscape can bring robustness~\cite{chen2020on,uday2019understanding,pu2020bridging}. For example, as formulated in Eq.~\ref{AT}, adversarial training (AT) enhances robustness via restricting the change of loss when the input of models is perturbed indeed. Thus, we want to theoretically justify why AT-SimGRACE works via validating that AT-SimGRACE can flatten the loss landscape. Inspired by previous work~\cite{neyshabur2017exploring} that connects sharpness of loss landscape and PAC-Bayes theory~\cite{mcallester1999some,mcallester1999pac-bayesian}, we utilize PAC-Bayes framework to derive guarantees on the expected error. Assuming that the prior distribution $P$ over the weights is a zero mean, $\sigma^{2}$ variance Gaussian distribution, with  
probability at least $1-\delta$ over the draw of $M$ graphs, the expected error of the encoder can be bounded as:
\begin{equation}
\label{TJ1}
\mathbb{E}_{\left\{\mathcal{G}_{i}\right\}_{i=1}^{M}, \boldsymbol{\Delta}}[\mathcal{L}(\boldsymbol{\theta}+\boldsymbol{\Delta})] \leq \mathbb{E}_{\boldsymbol{\Delta}}[\mathcal{L}(\boldsymbol{\theta}+\boldsymbol{\Delta})]+4 \sqrt{\frac{KL(\boldsymbol{\theta}+\boldsymbol{\Delta} \| P)+\ln \frac{2M}{\delta}}{M}}.
\end{equation}
We choose $\boldsymbol{\Delta}$ as a zero mean spherical Gaussian perturbation with variance $\sigma^{2}$ in every direction, and set the variance of the perturbation to the weight with respect to its magnitude $\sigma=\alpha\|\boldsymbol{\theta}\|$. Besides, we 
substitute $\mathbb{E}_{\boldsymbol{\Delta}}[\mathcal{L}(\boldsymbol{\theta}+\boldsymbol{\Delta})]$ with $\mathcal{L}(\boldsymbol{\theta}) + {\mathbb{E}_{\boldsymbol{\Delta}}[\mathcal{L}(\boldsymbol{\theta}+\boldsymbol{\Delta})]- \mathcal{L}(\boldsymbol{\theta})}$. Then, we can rewrite Eq.~\ref{TJ1} as:
\begin{equation}
\begin{aligned}
\label{TJ2}
\mathbb{E}_{\left\{\mathcal{G}_{i}\right\}_{i=1}^{M}, \boldsymbol{\Delta}}[\mathcal{L}(\boldsymbol{\theta}+\boldsymbol{\Delta})] \leq \mathcal{L}(\boldsymbol{\theta}) &+ \underbrace{\left\{\mathbb{E}_{\boldsymbol{\Delta}}[\mathcal{L}(\boldsymbol{\theta}+\boldsymbol{\Delta})]-\mathcal{L}(\boldsymbol{\theta})\right\}}_{\text{Expected sharpness}}\\
&+4 \sqrt{\frac{1}{M}\left(\frac{1}{2 \alpha}+\ln \frac{2M}{\delta}\right)}.
\end{aligned}
\end{equation}
It is obvious that $\mathbb{E}_{\boldsymbol{\Delta}}[\mathcal{L}(\boldsymbol{\theta}+\boldsymbol{\Delta})] \leq \max _{\boldsymbol{\Delta}}[\mathcal{L}(\boldsymbol{\theta}+\boldsymbol{\Delta})]$ and the third term $4 \sqrt{\frac{1}{M}\left(\frac{1}{2 \alpha}+\ln \frac{2M}{\delta}\right)}$ is a constant. Thus, AT-SimGRACE optimizes the worst-case of sharpness of loss landscape $\max_{\boldsymbol{\Delta}}[\mathcal{L}(\boldsymbol{\theta}+\boldsymbol{\Delta})]-\mathcal{L}(\boldsymbol{\theta})$ to the bound of the expected error, which explains why AT-SimGRACE can enhance the robustness.
\section{Experiments}
In this section, we conduct experiments to evaluate SimGRACE and AT-SimGRACE through answering the following research questions.
\begin{table*}[t]
\caption{Comparing classification accuracy with baselines under the same experiment setting. The top three accuracy or rank for each dataset are emphasized in bold. AR denotes average rank. $-$ indicates that results are not available in published papers.}
\label{Table 1}
\centering
\small
\resizebox{0.99\textwidth}{83.8pt}{
\begin{tabular}{c|cccc|cccc|c}
\hline \hline Methods & NCI1 & PROTEINS & DD & MUTAG & COLLAB & RDT-B & RDT-M5K & IMDB-B & AR $\downarrow$ \\
\hline \hline GL & $-$ & $-$ & $-$ & $81.66 \pm 2.11$ & $-$ & $77.34 \pm 0.18$ & $41.01 \pm 0.17$ & $65.87 \pm 0.98$ & $8.3$ \\
WL & \textbf{80.01} $\pm$ 0.50 & $72.92 \pm 0.56$ & $-$ & $80.72 \pm 3.00$ & $-$ & $68.82 \pm 0.41$ & $46.06 \pm 0.21$ & $\mathbf{72.30 \pm 3.44}$ & $6.2$ \\
DGK & \textbf{80.31} $\pm$ 0.46 & $73.30 \pm 0.82$ & $-$ & $87.44 \pm 2.72$ & $-$ & $78.04 \pm 0.39$ & $41.27 \pm 0.18$ & $66.96 \pm 0.56$ & $5.5$ \\
\hline node2vec & $54.89 \pm 1.61$ & $57.49 \pm 3.57$ & $-$ & $72.63 \pm 10.20$ & $-$ & $-$ & $-$ & $-$ & $9.0$ \\
sub2vec & $52.84 \pm 1.47$ & $53.03 \pm 5.55$ & $-$ & $61.05 \pm 15.80$ & $-$ & $71.48 \pm 0.41$ & $36.68 \pm 0.42$ & $55.26 \pm 1.54$ & $10.2$ \\
graph2vec & $73.22 \pm 1.81$ & $73.30 \pm 2.05$ & $-$ & $83.15 \pm 9.25$ & $-$ & $75.78 \pm 1.03$ & $47.86 \pm 0.26$ & $71.10 \pm 0.54$ & $6.7$ \\
MVGRL & $-$ & $-$ & $-$ & $75.40 \pm 7.80$ & $-$ & $82.00 \pm 1.10$ & $-$ & $63.60 \pm 4.20$ & $8.3$ \\
InfoGraph & $76.20 \pm 1.06$ & \textbf{74.44} $\pm$ 0.31 & $72.85 \pm 1.78$ & \textbf{89.01} $\pm$ 1.13 & \textbf{70.65} $\pm$ 1.13 & $82.50 \pm 1.42$ & $53.46 \pm 1.03$ & \textbf{73.03} $\pm$ 0.87 & $3.8$ \\
GraphCL & $77.87 \pm 0.41$ & $74.39 \pm 0.45$ & \textbf{78.62} $\pm$ 0.40 & $86.80 \pm 1.34$ & \textbf{71.36} $\pm$ 1.15 & \textbf{89.53} $\pm$ 0.84 & \textbf{55.99} $\pm$ 0.28 & \textbf{71.14} $\pm$ 0.44 & \textbf{3.1} \\
JOAO & $78.07 \pm 0.47$ & \textbf{74.55} $\pm$ 0.41 & $77.32 \pm 0.54$ & $87.35 \pm 1.02$ & $69.50 \pm 0.36$ & $85.29 \pm 1.35$ & $55.74 \pm 0.63$ & $70.21 \pm 3.08$ & $4.3$ \\
JOAOv2 & $78.36 \pm 0.53$ & $74.07 \pm 1.10$ & \textbf{77.40} $\pm$ 1.15 & \textbf{87.67} $\pm$ 0.79 & $69.33 \pm 0.34$ & \textbf{86.42} $\pm$ 1.45 & \textbf{56.03} $\pm$ 0.27 & $70.83 \pm 0.25$ & \textbf{3.6} \\
\hline \textbf{SimGRACE} & \textbf{79.12} $\pm$ 0.44 & \textbf{75.35} $\pm$ 0.09  & \textbf{77.44} $\pm$ 1.11 & \textbf{89.01} $\pm$ 1.31 & \textbf{71.72} $\pm$ 0.82 & \textbf{89.51} $\pm$ 0.89 & \textbf{55.91} $\pm$ 0.34 & \textbf{71.30} $\pm$ 0.77 & \textbf{2.0} \\
\hline \hline
\end{tabular}
}
\end{table*}
\begin{itemize}
\setlength{\itemsep}{2pt}
\setlength{\parsep}{2pt}
\setlength{\parskip}{2pt}
\item[$\bullet$] \textbf{RQ1. (Generalizability)} Does SimGRACE outperform competitors in unsupervised and semi-supervised settings?  
\item[$\bullet$] \textbf{RQ2. (Transferability)} Can GNNs pre-trained with SimGRACE show better transferability than competitors? 
\item[$\bullet$] \textbf{RQ3. (Robustness)} Can AT-SimGRACE perform better than existing competitors against various adversarial attacks?
\item[$\bullet$] \textbf{RQ4. (Efficiency)} How about the efficiency (time and memory) of SimGRACE? Does it more efficient than competitors?
\item[$\bullet$] \textbf{RQ5. (Hyperparameters Sensitivity)} Is the proposed SimGRACE sensitive to hyperparameters like the magnitude of the perturbation $\eta$, training epochs and batch size?
\end{itemize}
\subsection{Experimental Setup}

\subsubsection{Datasets.}  
For unsupervised and semi-supervised learning, we use datasets from the benchmark TUDataset~\cite{morris2020tudataset}, including graph data for various social networks~\cite{yanardag2015deep,benedek2020an} and biochemical molecules~\cite{riesen2008iam,dobson2003distinguishing}. For transfer learning, we perform pre-training on ZINC-2M and PPI-306K and finetune the model with various datasets including PPI, BBBP, ToxCast and SIDER.
\subsubsection{Evaluation Protocols.}
Following previous works for graph-level self-supervised representation learning~\cite{sun2019infograph,You2020GraphCL,you2021graph}, we evaluate the generalizability of the learned representations on both unsupervised and semi-supervised settings. In unsupervised setting, we train SimGRACE using the whole dataset to learn graph representations and feed them into a downstream SVM classifier with 10-fold cross-validation. For semi-supervised setting, we pre-train GNNs with SimGRACE on all the data and did finetuning \& evaluation with $K$ ($K=\frac{1}{\text { label rate }}$) folds for datasets without the explicit training/validation/test split. For datasets with the train/validation/test split, we pre-train GNNs with the training data, finetuning on the partial training data and evaluation on the validation/test sets. More details can be seen in the appendix.  

\subsubsection{Compared baselines.}
We compare SimGRACE with state-of-the-arts graph kernel methods including GL~\cite{shervashidze2009ecient}, WL~\cite{shervashidze2011weisfeiler-lehman} and DGK~\cite{yanardag2015deep}. Also, we compare SimGRACE with other graph self-supervised learning methods: GAE~\cite{kipf2016variational}, node2vec~\cite{grover2016node2vec}, sub2vec~\cite{adhikari2018sub2vec}, graph2vec~\cite{narayanan2017graph2vec}, EdgePred~\cite{hu2020strategies}, AttrMasking~\cite{hu2020strategies}, ContextPred~\cite{hu2020strategies}, Infomax (DGI)~\cite{velickovic2019deep}, InfoGraph~\cite{sun2019infograph} and instance-instance contrastive methods GraphCL~\cite{You2020GraphCL}, JOAO(v2)~\cite{you2021graph}. 
\subsection{Unsupervised and semi-supervised learning (RQ1)}

\begin{table*}[t]
\caption{Results for transfer learning setting. We report the mean (and standard deviation) ROC-AUC of 3 seeds with scaffold splitting. The top-3 accuracy for each dataset are emphasized in bold.}
\label{Table 4}
\setlength{\tabcolsep}{3pt}
\centering
\small
\fontsize{8.0pt}{\baselineskip}\selectfont
\begin{tabular}{c|c|ccccccccc}
\hline \hline
Pre-Train dataset & PPI-306K & \multicolumn{9}{c}{ZINC 2M}    \\ \hline
Fine-Tune dataset & PPI   & Tox21 & ToxCast & Sider & ClinTox & MUV & HIV  & BBBP & Bace & Average \\ \hline
No Pre-Train      & 64.8(1.0)  & 74.6 (0.4)& 61.7 (0.5) &58.2 (1.7) & 58.4 (6.4) & 70.7 (1.8) & 75.5 (0.8) &65.7 (3.3) & 72.4 (3.8) & 67.15  \\
EdgePred          & \textbf{65.7}(1.3)   &\textbf{76.0} (0.6)& 64.1 (0.6)& 60.4 (0.7)&64.1 (3.7) & 75.1 (1.2) &\textbf{76.3} (1.0) &67.3 (2.4) &\textbf{77.3} (3.5)  & 70.08\\
AttrMasking       & 65.2(1.6) & 75.1 (0.9) &\textbf{63.3} (0.6) &\textbf{60.5} (0.9)& 73.5 (4.3) &75.8 (1.0) &75.3 (1.5) &65.2 (1.4) &\textbf{77.8} (1.8) & 70.81  \\
ContextPred       & 64.4(1.3) & 73.6 (0.3)& 62.6 (0.6)&59.7 (1.8) &\textbf{74.0} (3.4) &72.5 (1.5) &\textbf{75.6} (1.0) &\textbf{70.6} (1.5) &\textbf{78.8} (1.2) & \textbf{70.93} \\
GraphCL           & \textbf{67.88}(0.85) &\textbf{75.1} (0.7)& \textbf{63.0} (0.4)& 59.8 (1.3)& \textbf{77.5} (3.8)& 76.4 (0.4)& 75.1 (0.7)& \textbf{67.8} (2.4)& 74.6 (2.1)& \textbf{71.16} \\

JOAO              & 64.43(1.38) & 74.8 (0.6)& 62.8 (0.7)&\textbf{60.4} (1.5) &66.6 (3.1) &\textbf{76.6} (1.7) &\textbf{76.9} (0.7) &66.4 (1.0) &73.2 (1.6) &69.71\\
\textbf{SimGRACE}   & \textbf{70.25}(1.22)  & \textbf{75.6} (0.5)&\textbf{63.4} (0.5) &\textbf{60.6} (1.0) &\textbf{75.6} (3.0) &\textbf{76.9} (1.3) &75.2 (0.9) & \textbf{71.3} (0.9)&75.0 (1.7) & \textbf{71.70} \\ \hline \hline
\end{tabular}
\end{table*}

\begin{table*}[t]
\caption{Comparing classification accuracy with baselines under the same semi-supervised setting. The top three accuracy or rank are emphasized in bold. $-$ indicates that label rate is too low for a given dataset size. LR and AR are short for label rate and average rank respectively.}
\label{Table 2}
\centering
\small
\resizebox{0.98\textwidth}{122.8pt}{
\begin{tabular}{c|c|ccc|ccc|c}
\hline \hline LR & Methods & NCI1 & PROTEINS & DD & COLLAB & RDT-B & RDT-M5K  & AR $\downarrow$ \\
\hline \hline  & No pre-train. & $60.72 \pm 0.45$ & $-$ & $-$ & $57.46 \pm 0.25$ & $-$ & $-$  & $8.5$ \\
& Augmentations & $60.49 \pm 0.46$ & $-$ & $-$ & $58.40 \pm 0.97$ & $-$ & $-$  & $8.0$ \\
\cline{2-8} & GAE & $61.63 \pm 0.84$ & $-$ & $-$ & $63.20 \pm 0.67$ & $-$ & $-$  & $5.5$ \\
& Infomax & \textbf{62.72} $\pm$ 0.65 & $-$ & $-$ & $61.70 \pm 0.77$ & $-$ & $-$  & $4.0$ \\
$1 \%$ & ContextPred & $61.21 \pm 0.77$ & $-$ & $-$ & $57.60 \pm 2.07$ & $-$ & $-$  & $7.5$ \\
& GraphCL & \textbf{62.55} $\pm$ 0.86 & $-$ & $-$ & \textbf{64.57} $\pm$ 1.15 & $-$ & $-$  & \textbf{2.0} \\
 & JOAO & $61.97 \pm 0.72$ & $-$ & $-$ & $63.71 \pm 0.84$ & $-$ & $-$  & $4.5$ \\
& JOAOv2 & $62.52 \pm 1.16$ & $-$ & $-$ & \textbf{64.51} $\pm$ 2.21 & $-$ & $-$  & \textbf{3.0} \\
\cline {2 - 8} & \textbf{SimGRACE} & \textbf{64.21} $\pm$ 0.65 & $-$ & $-$ & \textbf{64.28} $\pm$ 0.98 & $-$ & $-$  & \textbf{2.0} \\
\hline  & No pre-train. & $73.72 \pm 0.24$ & $70.40 \pm 1.54$ & $73.56 \pm 0.41$ & $73.71 \pm 0.27$ & $86.63 \pm 0.27$ & $51.33 \pm 0.44$  & $7.7$ \\
& Augmentations & $73.59 \pm 0.32$ & $70.29 \pm 0.64$ & $74.30 \pm 0.81$ & $74.19 \pm 0.13$ & $87.74 \pm 0.39$ & $52.01 \pm 0.20$  & $7.0$ \\
\cline { 2 - 8} & GAE & $74.36 \pm 0.24$ & $70.51 \pm 0.17$ & $74.54 \pm 0.68$ & \textbf{75.09} $\pm$ 0.19 & $87.69 \pm 0.40$ & $33.58 \pm 0.13$  & $6.3$ \\
& Infomax & \textbf{74.86}$\pm$ 0.26 & $72.27 \pm 0.40$ & $75.78 \pm 0.34$ & $73.76 \pm 0.29$ & $88.66 \pm 0.95$ & \textbf{53.61} $\pm$ 0.31 & $3.7$ \\
$10 \%$& ContextPred & $73.00 \pm 0.30$ & $70.23 \pm 0.63$ & $74.66 \pm 0.51$ & $73.69 \pm 0.37$ & $84.76 \pm 0.52$ & $51.23 \pm 0.84$  & $8.3$ \\
& GraphCL & \textbf{74.63}$\pm$ 0.25 &\textbf{74.17}$\pm$ 0.34 &\textbf{76.17}$\pm$ 1.37 &$74.23\pm 0.21$ &\textbf{89.11}$\pm$ 0.19 & $52.55\pm 0.45$ & \textbf{2.8} \\
& JOAO & $74.48\pm 0.27$ &$72.13\pm 0.92$ &$75.69\pm 0.67$ &\textbf{75.30} $\pm$ 0.32 &$88.14\pm 0.25$ &\textbf{52.83}$\pm$ 0.54 & $4.2$ \\
& JOAOv2 & \textbf{74.86}$\pm$ 0.39 &\textbf{73.31}$\pm$ 0.48 &\textbf{75.81}$\pm$ 0.73 &\textbf{75.53}$\pm$ 0.18 &\textbf{88.79}$\pm$ 0.65 &$52.71\pm 0.28$ & \textbf{2.5} \\ \cline {2-8}
& \textbf{SimGRACE} & $74.60 \pm 0.41$ & \textbf{74.03} $\pm$ 0.51 & \textbf{76.48} $\pm$ 0.52 & $74.74 \pm 0.28$ & \textbf{88.86} $\pm$ 0.62 & \textbf{53.97} $\pm$ 0.64 & \textbf{2.3} \\
\hline \hline
\end{tabular}
}
\end{table*}
\begin{table*}[t]
\caption{Performance under three adversarial attacks for GNN with different depth following the protocols in~\cite{dai2018adversarial}.}
\label{Table 5}
\centering
\small
\resizebox{0.98\textwidth}{40.0pt}{
\begin{tabular}{c|ccc|ccc|ccc}
\hline \hline \multirow{2}*{Methods} & \multicolumn{3}{|c|}{ Two-Layer } & \multicolumn{3}{c|}{ Three-Layer } & \multicolumn{3}{c}{ Four-Layer } \\
\cline {2 - 10} & No Pre-Train & GraphCL &\textbf{AT-SimGRACE} & No Pre-Train & GraphCL &\textbf{AT-SimGRACE} & No Pre-Train & GraphCL &\textbf{AT-SimGRACE} \\
\hline \hline Unattack & $93.20$ & \textbf{94.73} &$94.24$   & $98.20$ & $98.33$& \textbf{99.32}& $98.87$ & $99.00$ &$\textbf{99.13}$\\
\hline RandSampling & $78.73$ & $80.68$& \textbf{81.73} & $92.27$ & $92.60$& \textbf{94.27} & $95.13$ & 97.40& \textbf{97.67} \\
GradArgmax & $69.47$ & $69.26$ & \textbf{75.13} & $64.60$ & $89.33$ & \textbf{93.00}& $95.80$ & \textbf{97.00} & $96.60$\\
RL-S2V & $42.93$ & $42.20$& \textbf{44.86} & $41.93$ & $61.66$& \textbf{66.00}& $70.20$ & $84.86$& \textbf{85.29}\\
\hline \hline
\end{tabular}
}
\end{table*}
\begin{table}[t]
\caption{Comparisons of efficiency on three graph datasets. Note that we do not take the time for manual trial-and-errors of GraphCL into consideration. In fact, picking the suitable augmentations manually for GraphCL is much more time-consuming. All the three methods are evaluated on a 32GB V100 GPU.}
\label{Table 6}
\centering
\small
\resizebox{0.48\textwidth}{62.8pt}{
\begin{tabular}{cccc}\hline
\hline Dataset & Algorithm & Training Time & Memory \\
\hline & GraphCL & $111 s$    & $1231 MB$ \\
PROTEINS & JOAOv2 & $4088 s$  & $1403 MB$ \\
& \textbf{SimGRACE} & \textbf{46 s}    & \textbf{1175 MB} \\
\hline & GraphCL & $1033 s$   & $10199 MB$ \\
COLLAB & JOAOv2 & $10742 s$   & $7303 MB$ \\
        & \textbf{SimGRACE} & \textbf{378} s     & \textbf{6547 MB} \\
\hline & GraphCL & $917 s$   & $4135 MB$ \\
RDT-B & JOAOv2 & $10278 s$     & $3935 MB$ \\
    & \textbf{SimGRACE} & \textbf{280} s & \textbf{2729 MB} \\
\hline \hline
\end{tabular}
}
\end{table}

For unsupervised representation learning, as can be observed in Table~\ref{Table 1}, SimGRACE outperforms other baselines and always ranks top three on all the datasets. Generally, SimGRACE performs better on biochemical molecules compared with data augmentation based methods. The reason is that the semantics of molecular graphs are more fragile compared with social networks. General augmentations (drop nodes, drop edges and etc.) adopted in other baselines will not alter the semantics of social networks dramatically. 
For semi-supervised task, as can be observed in Table~\ref{Table 2}, we report two semi-supervised tasks with 1 \% and 10\% label rate respectively. In 1\% setting, SimGRACE outperforms previous baselines by a large margin or matching the performance of SOTA methods. For 10 \% setting, SimGRACE performs comparably to SOTA methods including GraphCL and JOAO(v2) whose augmentations are derived via expensive trial-and-errors or cumbersome search.  
\subsection{Transferability (RQ2)}
To evaluate the transferability of the pre-training scheme, we conduct experiments on transfer learning on molecular property prediction in chemistry and protein function prediction in biology following previous works~\cite{You2020GraphCL,hu2020strategies,Xia2022.02.03.479055}. Specifically, we pre-train and finetune the models with different datasets. For pre-training, learning rate is tuned in $\left\{0.01, 0.1, 1.0\right\}$ and epoch number in $\left\{20,40,60,80,100\right\}$ where grid serach is performed. As sketched in Table~\ref{Table 4}, there is no universally beneficial pre-training scheme especially for the out-of-distribution scenario in transfer learning. However, SimGRACE shows competitive or better transferability than other pre-training schemes, especially on PPI dataset.
\subsection{Adversarial robustness (RQ3)}
Following previous works~\cite{dai2018adversarial,You2020GraphCL}, we perform on synthetic data to classify the component number in graphs, facing the RandSampling, GradArgmax and RL-S2V attacks, to evaluate the robustness of AT-SimGRACE. To keep fair, we adopt Structure2vec~\cite{dai2016discriminative} as the GNN encoder as in~\cite{dai2018adversarial,You2020GraphCL}. Besides, we pretrain the GNN encoder for 150 epochs because it takes longer time for the convergence of adversarial training. We set the inner learning rate $\zeta=0.001$ and the radius of perturbation ball $\epsilon=0.01$. As demonstrated in Table~\ref{Table 5}, AT-SimGRACE boosts the robustness of GNNs dramatically compared with training from scratch and GraphCL under three typical evasion attacks. 
\subsection{Efficiency (Training time and memory cost) (RQ4)}
\begin{figure*}[t]
    \subfigure[NCI1]{
    \label{fig4-a}
    \includegraphics[width=0.24\textwidth]{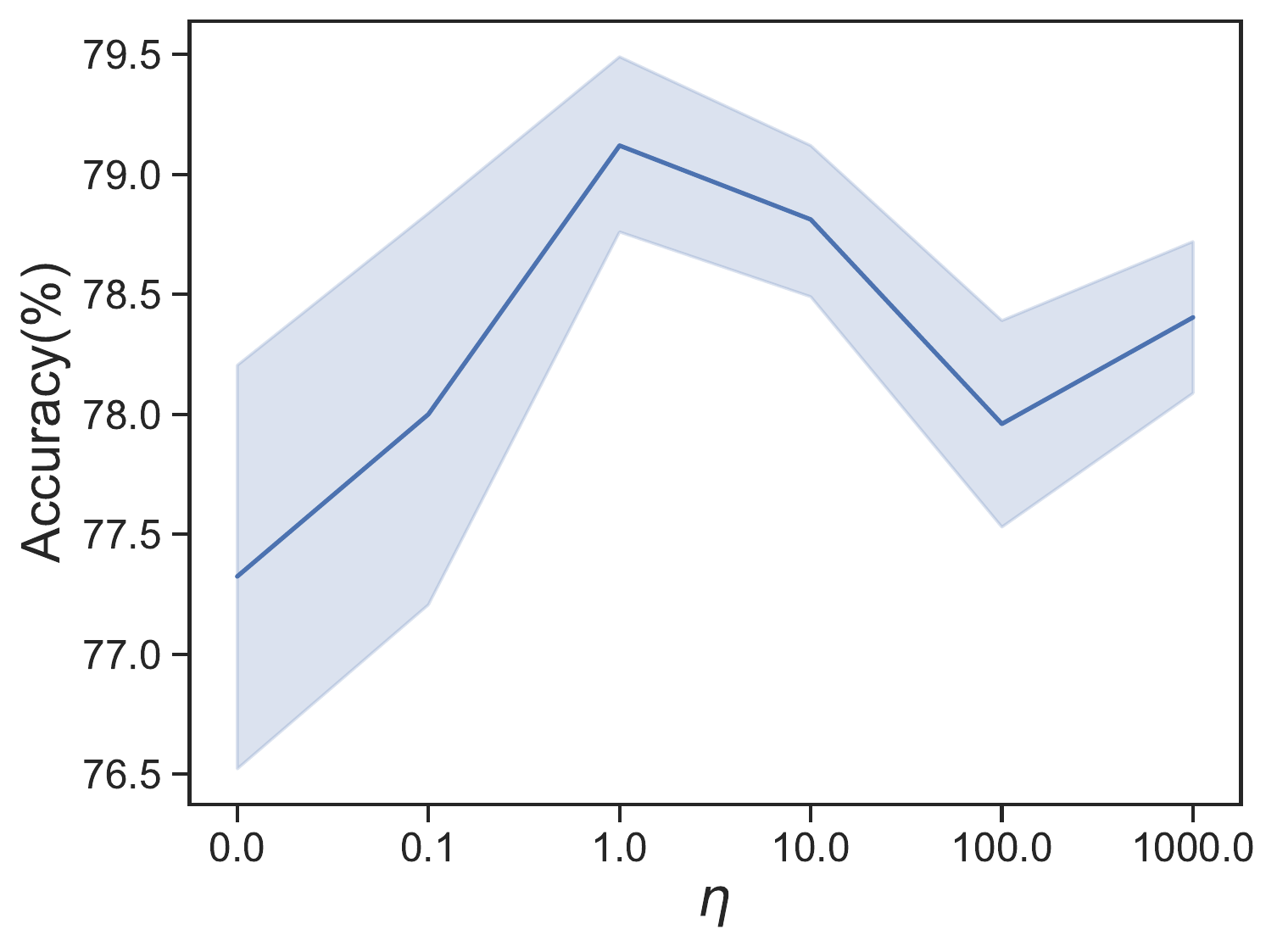}}
    \subfigure[MUTAG]{
    \label{fig4-b}
    \includegraphics[width=0.24\textwidth]{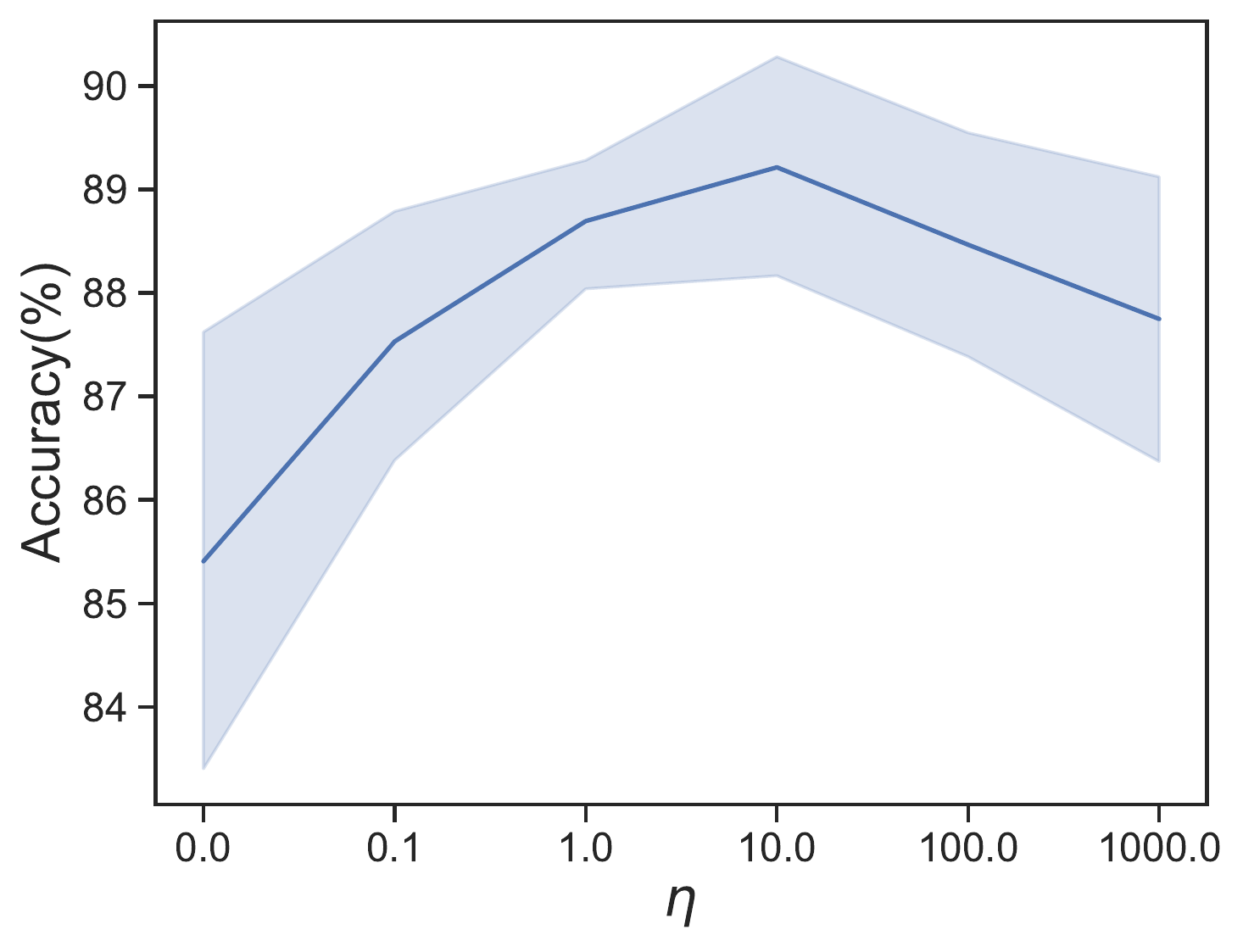}}
    \subfigure[COLLAB]{
    \label{fig4-c}
    \includegraphics[width=0.24\textwidth]{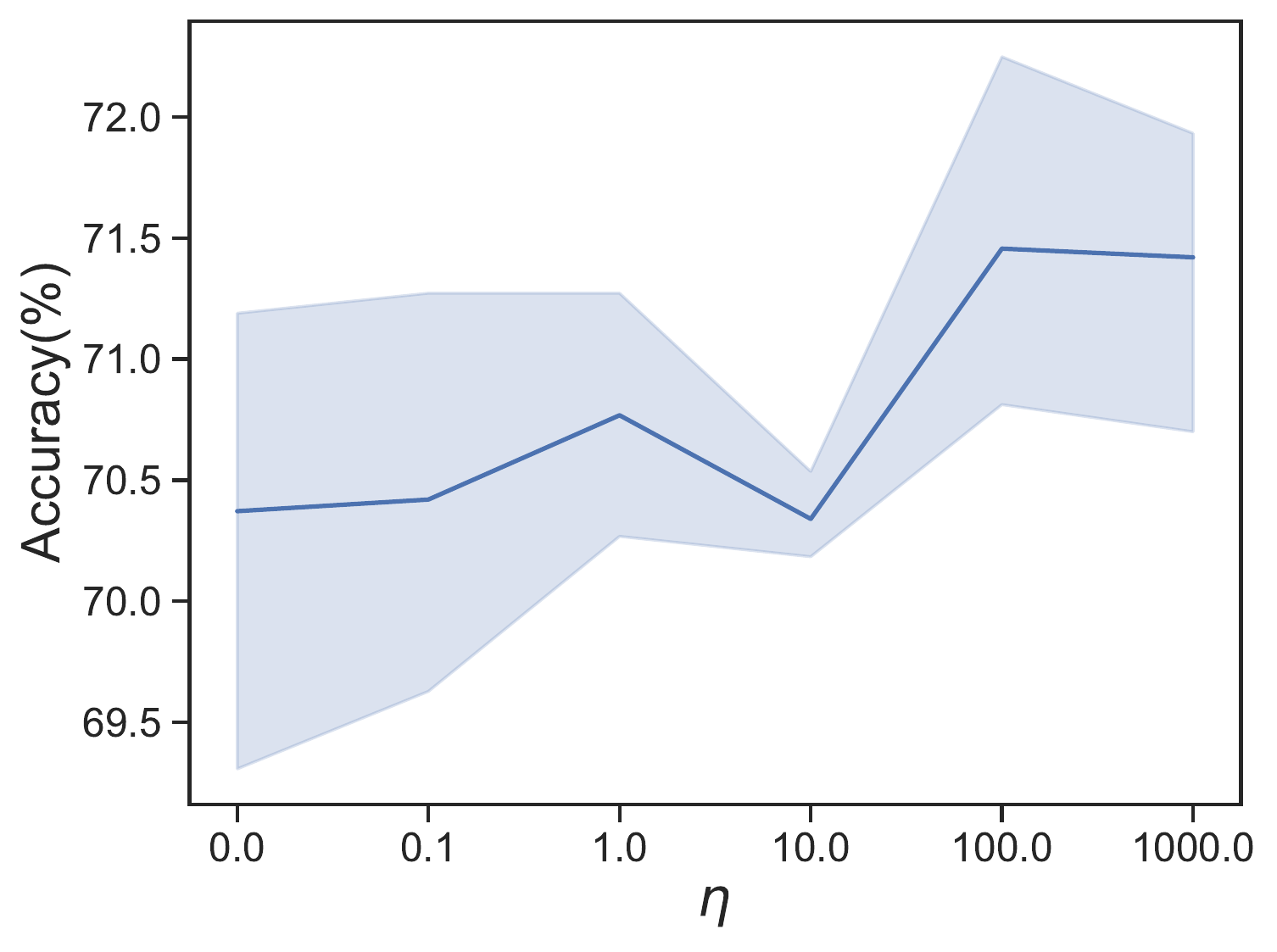}}
    \subfigure[RDT-5K]{
    \label{fig4-d}
    \includegraphics[width=0.24\textwidth]{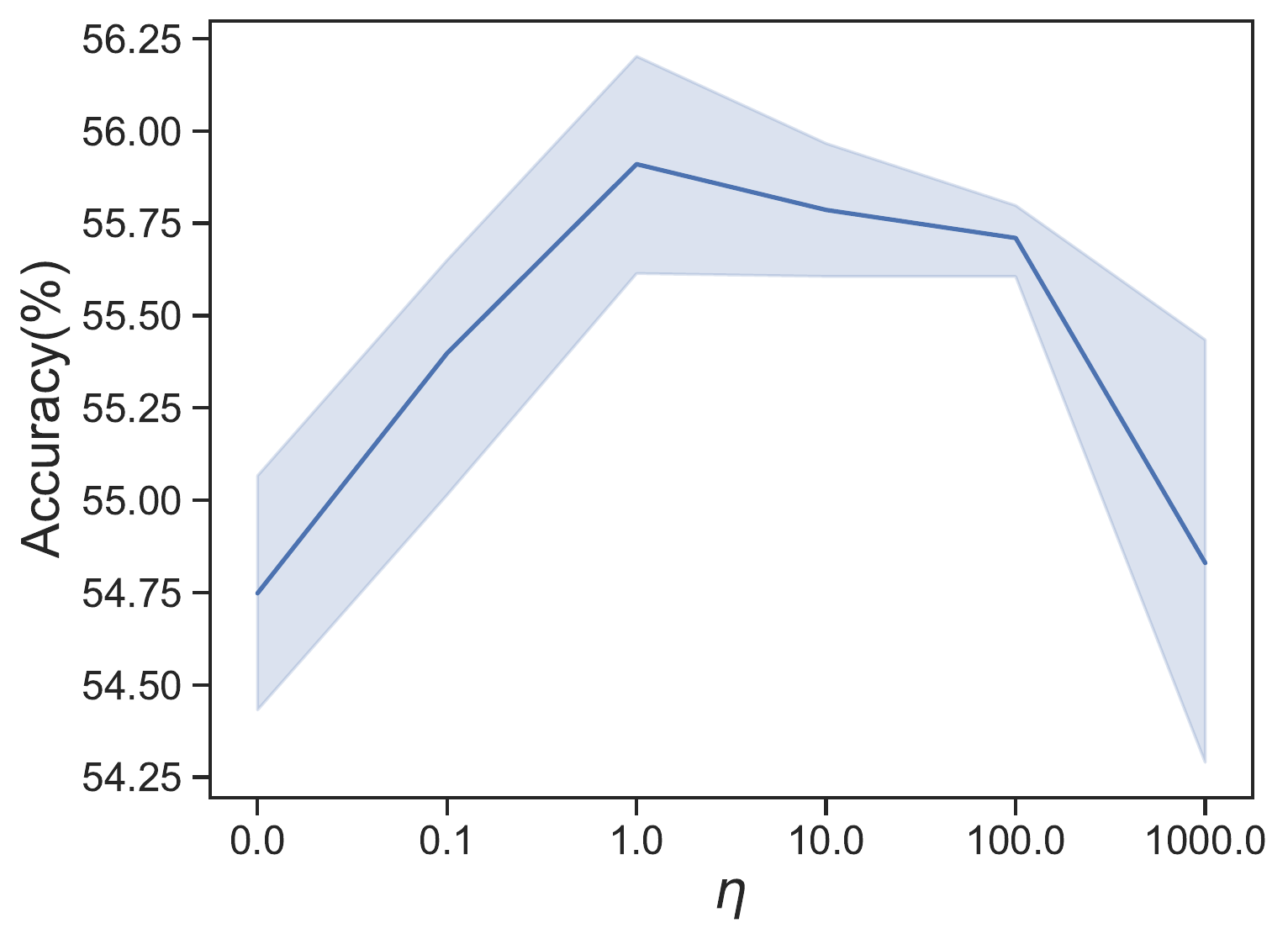}}
    \caption{Performance versus magnitude of the perturbation ($\eta$) in unsupervised representation learning task.}
    \label{fig4}
\end{figure*}
In Table~\ref{Table 6}, we compare the performance of SimGRACE with the state-of-the-arts methods including GraphCL and JOAOv2 in terms of their training time and the memory overhead. Here, the training time refers to the time for pre-training stage of the semi-supervised task and the memory overhead refers to total memory costs of model parameters and all hidden representations of a batch. As can be observed, SimGRACE runs near 40-90 times faster than JOAOv2 and 2.5-4 times faster than GraphCL. If we take the time for manual trial-and-errors in GraphCL into consideration, the superiority of SimGRACE will be more 
pronounced. Also, SimGRACE requires less computational memory than GraphCL and JOAOv2. In particular, the efficiency of SimGRACE can be more prominent on large-scale social graphs, such as COLLAB and RDT-B.  
\subsection{Hyper-parameters sensitivity analysis (RQ5)}
\subsubsection{Magnitude of the perturbation}
\label{HP}
As can be observed in Figure~\ref{fig4}, weight perturbation is crucial in SimGRACE. If we set the magnitude of the perturbation as zero ($\eta=0$), the performance is usually the lowest compared with other setting of perturbation across these four datasets. This observation aligns with our intuition. Without perturbation, SimGRACE simply compares two original samples as a negative pair while the positive pair loss becomes zero, leading to homogeneously pushes all graph representations away from each other, which is non-intuitive to justify. Instead, appropriate perturbations enforce the model to learn representations invariant to the perturbations through maximizing the agreement between a graph and its perturbation. Besides, well aligned with previous works~\cite{robinson2021contrastive,ho2020contrastive} that claim "hard" positive pairs and negative pairs can boost the performance of contrastive learning, we can observe that larger magnitude (within an appropriate range) of the perturbation can bring consistent improvement of the performance. However, over-large perturbations will lead to performance degradation because the semantics of graph data are not preserved.
\subsubsection{Batch-size and training epochs}
\begin{figure}[ht]
    \begin{center}
    \includegraphics[width=0.45\textwidth]{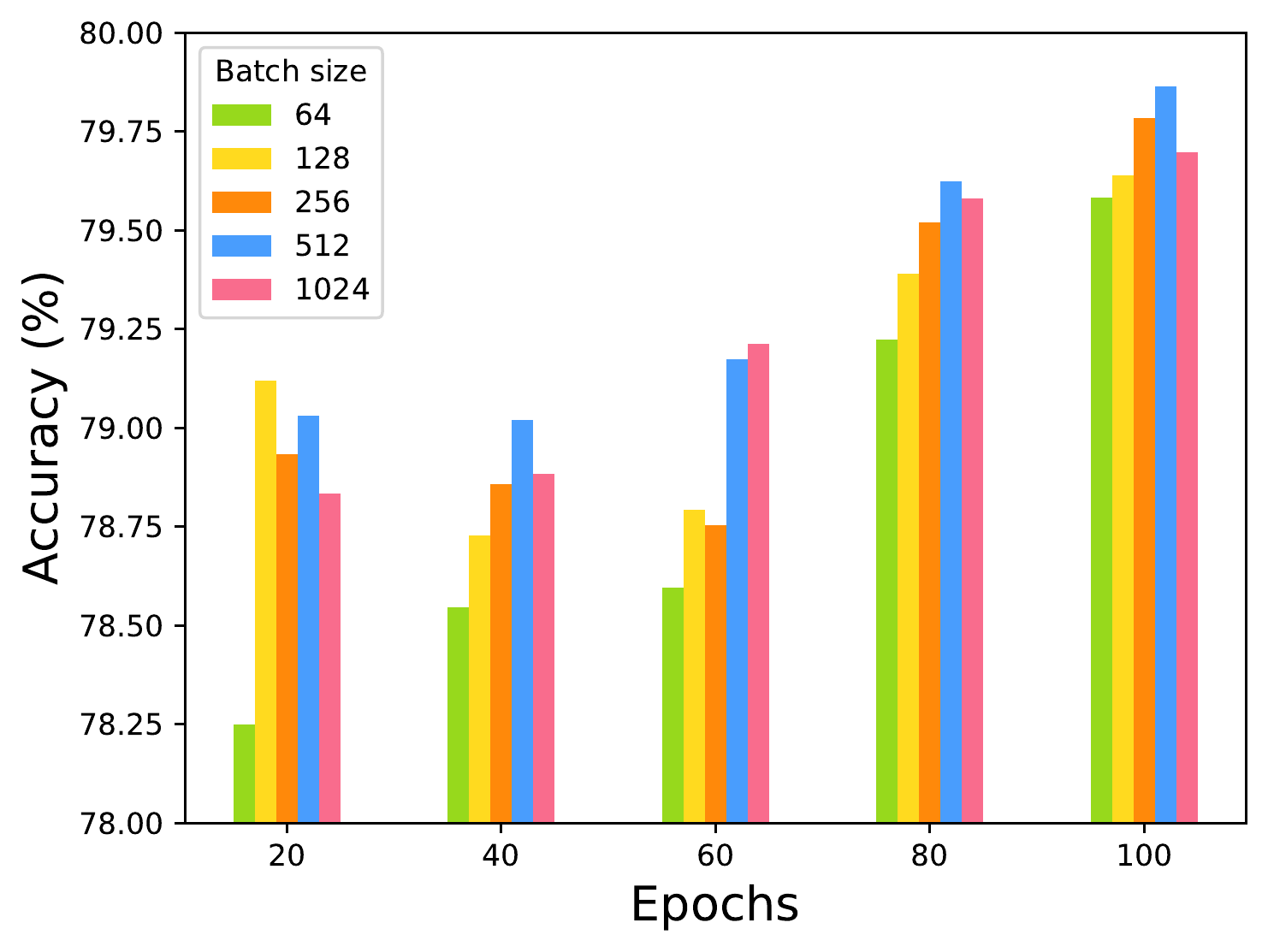}
    \end{center}
    \caption{Performance of SimGRACE trained with different batch size and epochs on NCI1 dataset.}
    \label{fig_4}
\end{figure}
Figure~\ref{fig_4} demonstrates the performance of SimGRACE trained with various batch size and epochs. Generally, larger batch size or training epochs can bring better performance. The reason is that larger batch size will provide more negative samples for contrasting. Similarly, training longer also provides more new negative samples for each sample because the split of total datasets is more various with more training epochs. In our experiments, to keep fair, we follow the same settings of other competitors~\cite{You2020GraphCL,you2021graph} via training the GNN encoder with batch size as 128 and number of epochs as 20. In fact, we can further improve the performance of SimGRACE with larger batch size and longer training time. 
\section{Conclusions}
In this paper, we propose a simple framework (SimGRACE) for graph contrastive learning. Although it may appear simple, we demonstrate that SimGRACE can outperform or match the state-of-the-art competitors on multiple graph datasets of various scales and types, while enjoying unprecedented degree of flexibility, high efficiency and ease of use. We emancipate graph contrastive learning from tedious manual tuning, cumbersome search or expensive domain knowledge. Furthermore, we devise adversarial training schemes to enhance the robustness of SimGRACE in a principled way and theoretically explain the reasons. There are two promising avenues for future work: 
(1) exploring if encoder perturbation can work well in other domains like computer vision and natural language processing. 
(2) applying the pre-trained GNNs to more real-world tasks including social analysis and biochemistry.
\begin{acks}
This work is supported in part by the Science and Technology Innovation 2030 - Major Project (No. 2021ZD0150100) and National Natural Science Foundation of China (No. U21A20427).
\end{acks}
\bibliographystyle{ACM-Reference-Format}
\bibliography{main}

\newpage
\appendix
\section{Appendix: Datasets in Various Settings}
\subsection{Unsupervised learning \& Semi-supervised learning}
\begin{table}[ht]
\caption{Datasets statistics for unsupervised and semi-supervised experiments.}
\label{table 1}
\centering
\small
\resizebox{0.48\textwidth}{46pt}{
\begin{tabular}{c|c|c|c|c}
\hline \hline Datasets & Category & Graph Num. & Avg. Node & Avg. Degree \\
\hline \hline NCI1 & Biochemical Molecules & 4110 & $29.87$ & $1.08$ \\
PROTEINS & Biochemical Molecules & 1113 & $39.06$ & $1.86$ \\
DD & Biochemical Molecules & 1178 & $284.32$ & $715.66$ \\
MUTAG & Biochemical Molecules & 188 & $17.93$ & $19.79$ \\
\hline COLLAB & Social Networks & 5000 & $74.49$ & $32.99$ \\
RDT-B & Social Networks & 2000 & $429.63$ & $1.15$ \\
RDB-M & Social Networks & 2000 & $429.63$ & $497.75$ \\
IMDB-B & Social Networks & 1000 & $19.77$ & $96.53$ \\
\hline \hline
\end{tabular}
}
\end{table}
For unsupervised setting, experiments are performed for 5 times each of which corresponds to a 10-fold evaluation, with mean and standard deviation of accuracies (\%) reported. For semi-supervised learning, we perform experiments with 1\% (if there are over 10 samples for each class) and 10\% label rate for 5 times, each of which corresponds to a 10-fold evaluation, with mean and standard deviation of accuracies (\%) reported. For pre-training, learning rate is tuned in $\left\{0.1, 1.0, 5.0, 10.0\right\}$ and epoch number in $\left\{20, 40, 60, 80, 100\right\}$ where grid search is performed. All datasets used in both unsupervised and semi-supervised experiments can be seen in Table~\ref{table 1}.
\subsection{Transfer learning}
\begin{table}[ht]
\caption{Datasets statistics for transfer learning.}
\label{table 2}
\centering
\small
\resizebox{0.488\textwidth}{28pt}{
\begin{tabular}{c|c|c|c|c|c}
\hline \hline Datasets & Category & Utilization & Graph Num. & Avg. Node & Avg. Degree \\
\hline \hline ZINC-2M & Biochemical Molecules & Pre-Training & 2,000,000 & $26.62$ & $57.72$ \\
PPI-306K & Protein-Protein Intersection Networks & Pre-Training & 306,925 & $39.82$ & $729.62$ \\
\hline BBBP & Biochemical Molecules & Finetuning & 2,039 & $24.06$ & $51.90$ \\
ToxCast & Biochemical Molecules & Finetuning & 8,576 & $18.78$ & $38.52$ \\
SIDER & Biochemical Molecules & Finetuning & 1,427 & $33.64$ & $70.71$ \\
\hline\hline
\end{tabular}}
\end{table}
The datasets utilized in transfer learning can be seen in Table~\ref{table 2}. ZINC-2M and PPI-306K are used for pre-training and the left ones are for fine-tuning.
\section{GNN architectures in Various Settings}
 To keep fair, we adopt the same GNNs architectures with previous competitors. Specifically, for unsupervised task, GIN~\cite{xu2019how} with 3 layers and 32 hidden dimensions is adopted as the encoder. For semi-supervised task, we utilize ResGCN~\cite{chen2019are} with 5 layers and 128 hidden dimensions. For transfer learning, we adopt GIN with the default setting in~\cite{hu2020strategies} as the GNN-based encoder. For experiments on adversarial robustness, Structure2vec is adopted as the GNN-based encoder as in~\cite{dai2018adversarial}.
\end{document}